\documentclass{article}
\usepackage[preprint]{log_2026}			

\usepackage{booktabs}						
\usepackage{multirow}						
\usepackage{amsfonts}						
\usepackage{graphicx}						
\usepackage{duckuments}						
\usepackage{subcaption}
\usepackage[numbers,compress,sort]{natbib}	

\usepackage{wrapfig}

\title[Train Small, Deploy Large: Zero-Shot GNN Transfer Through Geometric Renormalization]{Train Small, Deploy Large: \\Zero-Shot GNN Transfer Through Geometric Renormalization}

\author[Jankowski et al.]{%
Robert Jankowski\\
TU Delft \\
\email{R.Jankowski@tudelft.nl}\And
Pedro Almagro-Blanco\\
University of Seville\\
\email{palmagro@us.es}\And
Mari\'an Bogu\~n\'a\\
University of Barcelona\\
\email{marian.boguna@ub.edu}\And
Melanie Weber\\
Harvard University\\
\email{mweber@seas.harvard.edu}\And
M. Ángeles Serrano\\
University of Barcelona and ICREA\\
\email{marian.serrano@ub.edu}
}

\begin{document}

\maketitle

\begin{abstract}
Graph neural networks (GNNs) can operate on large graphs but become infrastructure-sensitive at the scale of millions of nodes and typically require scalable training techniques for even larger graphs. This raises a central question: when can a model trained on a smaller, scaled-down replica of a graph be deployed on the full-resolution graph without retraining? We introduce a zero-shot transfer protocol in which a GNN is trained on a graph coarse-grained by geometric renormalization (GR), and the resulting weights are transferred directly to the original network. Across synthetic and real-world networks, training on GR scaled-down replicas preserves much of the original-scale predictive performance while significantly reducing training cost. We further find that learned representations and predictive trajectories remain aligned across scales. These findings suggest that structural similarity may be more important than network size in determining GNN transferability, opening a path toward scale-equivariant graph architectures.
\end{abstract}

\begin{figure}[h]
	\centering	
    \includegraphics[width=\linewidth]{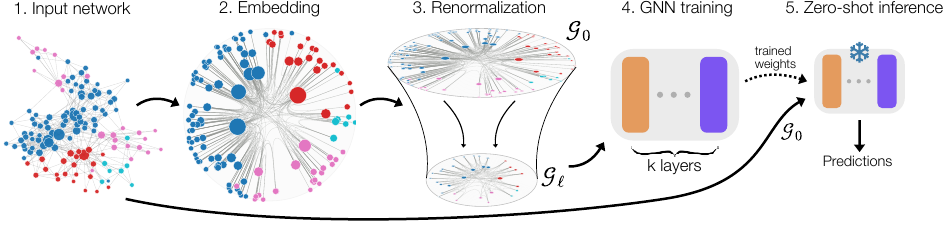}
	\caption{\textbf{Overview of the train-small, deploy-large protocol}. The input graph $G_0$ is embedded and renormalized to obtain a scaled-down graph $G_\ell$. A GNN is trained on $G_\ell$, and the learned weights are transferred without retraining to $G_0$ for zero-shot inference.}
	\label{fig:pipeline}
\end{figure}

\section{Introduction}
Message-passing GNNs learn node, edge, and graph representations by iteratively propagating and aggregating information among adjacent nodes of the input graph~\cite{kipf2017semisupervised,hamilton2017}. Although a given GNN architecture can, in principle, be applied to graphs of any size, successful transfer across graph scales is not guaranteed. A GNN trained at one scale may perform poorly at another when the structural properties encountered during training differ from those present at deployment~\cite{yehudai2021local}. The central challenge is therefore not merely computational scalability, but understanding how changes in graph resolution induce distribution shifts that can alter the statistics governing message passing and compromise transfer across scales.

Addressing this challenge requires identifying the structural conditions under which GNN parameters learned on a scaled-down replica of a graph can be transferred to the full resolution graph without retraining. Existing results establish GNN transferability across graphs under restrictive assumptions, such as dense graphon limits~\cite{ruiz2020graphon}. However, real networks are often sparse, heterogeneous, and clustered~\cite{newman2003structure}, and standard message-passing GNNs can yield markedly different representations of the same network at different resolutions or levels of coarse-graining~\cite{koke2026graph}. These observations suggest that cross-scale GNN transfer requires more than size-independent weights. It requires a scale transformation that preserves the statistics used by message passing.

Graph coarsening provides one route to smaller training graphs, but most existing methods are designed to preserve global properties rather than the local neighborhood statistics that govern message passing in GNNs~\cite{loukas2019graph,chen2022graph}. As a result, a coarsened graph may remain a good approximation of the original in the spectral sense while substantially altering degree distributions, clustering spectra, or community structure. Recent work has derived message-passing guarantees for graph coarsening, requiring a purpose-built propagation operator and approximate original-graph training~\cite{joly2024graph}.

Our approach builds on a different principle: many real networks are 
shaped by an underlying latent space~\cite{serrano2008self,boguna2021network} with hyperbolic geometry~\cite{krioukov2010hyperbolic} in which pairwise distances between nodes determine the likelihood of connection. The geometric framework explains characteristic network properties such as the small-world property, heterogeneous degree distributions, high clustering, and self-similar organization~\cite{serrano2008self,serrano2022shortest}. Geometric renormalization (GR)~\cite{garcia2018multiscale} is a technique that merges nearby nodes along the similarity subspace, producing a hierarchy of graph representations at progressively 
lower resolutions that preserve the multiscale organization of the original network. Therefore, GR is not merely a graph compression procedure but a geometry-preserving scale transformation.

In this work, we study zero-shot GNN weight transfer across network resolutions defined by a geometric renormalization flow. Given an original graph $G_0$ and a renormalized graph $G_\ell$, we train a GNN on $G_\ell$ and deploy the same weights directly on $G_0$. This train-small, deploy-large protocol tests whether the learned parameters are approximately invariant across network scales. We evaluate transfer using node classification accuracy, similarity of learned representations~\cite{kornblith2019similarity}, and alignment of predictive trajectories during training. On real networks and synthetic networks of the HypBench framework~\cite{Aliakbarisani2026}, generated using the $\mathbb{S}^1/\mathbb{H}^2$ model, we find that predictive performance can be largely preserved under zero-shot transfer as GR preserves the relevant geometric and topological structure.

Our contributions are five-fold:
\begin{itemize}
    \item We introduce a train-small, deploy-large protocol for studying zero-shot GNN weight transfer across graph resolutions.
    \item We provide empirical evidence that GR preserves GNN transfer performance under substantial graph compression on synthetic and real networks.
    \item We show that models trained at different GR scales learn compatible node representations, as quantified by centered kernel alignment (CKA) and orthogonal Procrustes (OP) similarity measures.
    \item We demonstrate that GR preserves functional training trajectories across graph scales, as quantified by the Jensen-Shannon divergence between output-probability distributions.
    \item We release \textit{cuMercator}, a GPU implementation of Mercator that achieves a speedup of more than 400$\times$ in computing hyperbolic embeddings for networks with 10,000 nodes, with even greater speedups for larger networks; this implementation may be of independent interest.
\end{itemize}

\section{Related work} 

In machine learning, graph coarsening is widely used to reduce graph size or construct hierarchical representations. Learned pooling methods, such as DiffPool~\cite{ying2018hierarchical} and Graph U-Nets~\cite{gao2019graph}, build task-dependent coarse graphs inside the model. This line of work is closely related to graph pooling, where coarse representations are learned through node assignments, edge contractions, or geometric reductions~\cite{grattarola2022understanding,limbeck2026geometry,fenggraph}. Preprocessing approaches instead coarsen the input graph before training; for example, Huang et al.~\cite{huang2021scaling} used graph coarsening to scale GNN training. More recently, Joly and Keriven~\cite{joly2024graph} argued that classical spectral coarsening does not automatically preserve message passing and proposed coarsening operators with message-passing guarantees. These works primarily focus on efficient training, hierarchical pooling, or signal preservation on the coarse graph. 

In network science, network renormalization provides a framework for relating graph structure across scales~\cite{gabrielli2025network}. Early work by Song, Havlin, and Makse~\cite{song2005self} introduced box-covering renormalization and showed that many real networks exhibit self-similar organization. Hidden-metric-space models later connected network self-similarity to latent geometry~\cite{serrano2008self,krioukov2010hyperbolic}. Geometric renormalization builds on this idea by embedding a graph in the hyperbolic plane and merging nearby nodes in the similarity subspace, producing a multiscale flow that preserves the probability of connection and therefore key topological observables~\cite{garcia2018multiscale}. This framework has been applied to empirical systems such as the human connectome~\cite{zheng2020geometric} and extended to weighted networks~\cite{zheng2024geometric}. Other renormalization schemes define scales without hidden geometry, including Laplacian renormalization for heterogeneous networks~\cite{villegas2023laplacian}, higher-order Laplacian renormalization for simplicial and hypergraph structures~\cite{nurisso2025higher}, and geometry-free multiscale renormalization based on hidden variables~\cite{garuccio2023multiscale}. Among these methods, GR is distinguished by its direct connection to analytically tractable hidden-metric-space models, which makes its coarse-graining and rescaling rules interpretable as a renormalization flow in the underlying similarity subspace rather than merely as an algorithmic graph contraction. These approaches define principled graph-scale transformations, but have not been used to study the scale dependence of learned GNN weights.

GNNs can be applied to graphs of different sizes because their parameters are shared across neighborhoods. However, this architectural size agnosticism does not guarantee cross-size generalization. More broadly, GNN performance has been shown to depend sensitively on graph-specific data properties, including degree distributions, joint topological and feature structure, and spectral characteristics~\citep{li2023metadata,fesser2025performance}. Yehudai et al.~\cite{yehudai2021local} showed that GNNs may fail to generalize from small to large graphs when local structures differ across sizes. Graphon-based analyses provide transferability guarantees for graph sequences sampled from a common limiting object~\cite{ruiz2020graphon,ruiz2023transferability}, while recent work shows that standard message-passing GNNs can be discontinuous across graph resolutions~\cite{koke2026graph}. These results demonstrate that GNN transfer across scales is nontrivial.

\section{Network geometry and geometric renormalization}\label{sec:GR}

Random geometric graphs describe complex networks by embedding nodes in a latent metric space, where geometric distances encode similarity and link formation emerges from a trade-off between node popularity and similarity~\cite{serrano2008self}, with emerging hyperbolic geometry~\cite{krioukov2010hyperbolic}. 
The canonical model in this framework is the $\mathbb{S}^1$ model~\cite{serrano2008self}, which places nodes on a circle of radius $R=N/2\pi$, representing the similarity space, and connects every pair $i$, $j$ with probability
\begin{align}\label{eq:s1}
    p_{ij} = \frac{1}{1 + \chi^\beta} = \frac{1}{1 + \left(\frac{R\Delta\theta_{ij}}{\mu\kappa_i\kappa_j}\right)^\beta},
\end{align}
where $\mu$ controls the average degree of the network, $\beta$ its clustering coefficient, and $\Delta\theta_{ij}$ is the angular distance between nodes $i$ and $j$. The hidden degrees $\kappa$ reflect node popularity or importance. The $\mathbb{S}^1$ model is isomorphic to the purely geometric $\mathbb{H}^2$ model~\cite{krioukov2010hyperbolic}, in which nodes are placed in a two-dimensional hyperbolic disk of radius $R_{\mathbb{H}^2} = 2\ln\left(\frac{2R}{\mu \kappa_0^2}\right)$, where $\kappa_0 = \min \{\kappa_i\}$. By mapping each $\kappa_i$ to a radial coordinate $r_i$, 
$r_i = R_{\mathbb{H}^2} - 2\ln \frac{\kappa_i}{\kappa_0}$, the connection probability, Eq.~\ref{eq:s1}, becomes
\begin{align}
    p_{ij} = \frac{1}{1 + e^{\frac{\beta}{2}(x_{ij} - R_{\mathbb{H}^2})}},
\end{align}
where $x_{ij} = r_i + r_j + 2\ln (\Delta\theta_{ij}/2)$ is a good approximation of the hyperbolic distance between two points with coordinates $(r_i, \theta_i)$ and $(r_j, \theta_j)$ in the native representation of hyperbolic space~\cite{serrano2022shortest}.

\begin{figure}[h]
	\centering	
    \includegraphics[width=\linewidth]{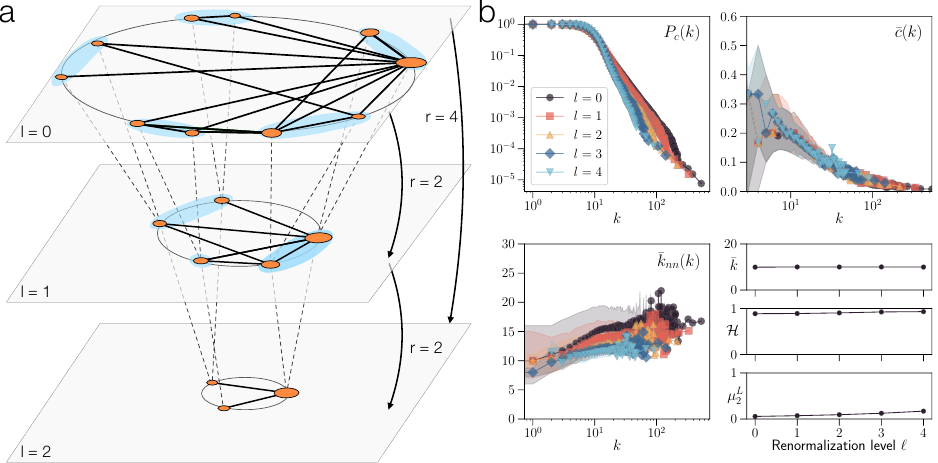}
	\caption{(\textbf{a}) Schematic of geometric renormalization. Each layer is obtained by applying a renormalization step with resolution $r$, starting from the original network at $l=0$. Orange nodes are placed on the circle, with sizes proportional to the logarithm of their hidden degrees, and solid lines denote links within each layer. Blue shaded regions indicate coarse-graining blocks, and dashed lines connect nodes to their supernodes in layer $\ell+1$. Two supernodes are linked in layer $\ell+1$ if at least one pair of nodes belonging to them is linked in layer $\ell$. Under composition, GR forms an Abelian semigroup: repeated transformations at a given resolution are equivalent to a single transformation at higher resolution. If the number of nodes is not divisible by $r$, the final supernode contains fewer than $r$ nodes, as shown at $\ell = 1$. Reproduced from~\cite{garcia2018multiscale}. (\textbf{b}) Topological validation of GR for synthetic networks in the small-world regime. In the top-left panel, the complementary cumulative degree distribution $P_c(k)$ is shown across renormalization levels $l=0$ to $l=4$. The top-right panel shows the clustering spectrum $\bar c(k)$, and the bottom-left panel shows the average nearest-neighbor degree $\bar k_{nn}(k)$. In both panels, solid lines denote the median and shaded regions denote the interquartile range (IQR). The three panels on the bottom right show, as a function of 
    scale $l$, the mean degree $\bar k$, the homophily ratio $\mathcal{H}$, and the spectral gap $\mu_2^L$.}
	\label{fig:GR}
\end{figure}

Geometric renormalization (GR) unveils the self-similarity of complex networks to construct scaled-down representations~\cite{garcia2018multiscale}. It operates by merging neighboring nodes in the similarity subspace into supernodes and rescaling connections, averaging over short-range interactions and progressively preserving longer-range connections. The procedure is shown in Figure~\ref{fig:GR}a. Given an input network $G_0$ at layer 0, we first construct its hyperbolic map using the embedder Mercator~\cite{garcia2019mercator,jankowski2023d}, obtaining the hidden degree (or radial position) and angular position of each node, $(\kappa_i, \theta_i)$. Then, we partition the nodes into nonoverlapping blocks of $r$ consecutive nodes along the circle and coarse-grain each block into a supernode, independently of whether the nodes within the block are connected. Each supernode is placed within the angular region spanned by its corresponding block, thereby preserving the circular order of nodes in the original embedding throughout the renormalization process. Finally, all links between nodes belonging to two different supernodes, if any, are 
rescaled into a single link between those supernodes. The GR transformation generates a version of the original network with the same statistical properties while typically increasing the average degree. The additional links are then selectively pruned to preserve the network’s structure, yielding a scaled-down replica of the original network. In Appendix~\ref{apx:GR}, we provide a detailed description of the algorithm.

\section{Methods}

\subsection{Experimental setup}

In this work, we use synthetic networks generated with HypBench~\cite{Aliakbarisani2026}, together with a set of real networks. The schematic pipeline is shown in Figure~\ref{fig:pipeline}. Starting from the input graph $G_0$, we first embed the network into the hyperbolic disk using Mercator. For synthetic networks, this embedding step is not required because the ground-truth coordinates are known. We then apply GR to construct a coarse-grained network $G_\ell$, which is used for GNN training.
After training the GNN on $G_\ell$ for node classification, we evaluate the model directly on $G_0$ without any retraining: node representations are computed on $G_0$ using the weights learned on $G_\ell$, and predictions are obtained for every node in the original test set. We refer to the resulting accuracy, computed on the original test nodes of $G_0$, as the \emph{transfer test accuracy}.

Results are averaged over 10 random train/test splits per network (60/40 ratio), and, for synthetic networks, over 10 independently generated network realizations per configuration. 

To generate scale-down replicas of a given input network, we use the geometric renormalization procedure based on the family of hyperbolic random graph models $\mathbb{S}^1/\mathbb{H}^2$ as described in Section~\ref{sec:GR}. Because nodes are divided into train and test sets, the merging procedure must avoid label leakage. Each supernode label is assigned by majority vote over the training labels within the block; test labels are ignored. In the event of a tie, one of the tied labels is selected at random. This yields a fixed compression factor of $r$ per level. Unless stated otherwise, we use $r=2$ in all experiments. The feature vector of each supernode is obtained by averaging the feature vectors of its constituent nodes.

\paragraph{Random Baselines.}
As a baseline, we also consider random renormalization, in which both the network topology and the node coordinates are randomized before renormalization (\textit{Random}). To disentangle the contribution of each of these two factors, we additionally consider two partial baselines in which only the topology (\textit{Random-T}) or only the geometry (\textit{Random-G}) is randomized, while the other is kept unaltered, following the same topology/geometry disentanglement strategy used by Zheng et al.~\cite{zheng2020geometric}. In the main text, we report the results for GR against the fully randomized baseline (\textit{Random}). Results for \textit{Random-T} and \textit{Random-G} are reported in Appendix~\ref{apx:random_variants}.

\paragraph{Parameters of HypBench.}
The HypBench framework allows us to generate synthetic networks with nodes' features by varying parameters of the $\mathbb{S}^1/\mathbb{H}^2$ and bipartite-$\mathbb{S}^1/\mathbb{H}^2$ models~\cite{Aliakbarisani2026}. The unipartite networks are generated with $N=2^{17}=131{,}072$ nodes, $\langle k \rangle =10$ and $\beta=1.5$, $N_c=6$ classes, and homophily parameter $\alpha=6$. We consider two degree-distribution exponents, $\gamma=2.5$ and $\gamma=3.5$, corresponding to the ultra-small-world (USW) and small-world (SW) regimes, respectively~\cite{boguna2020smallworlds}. These regimes differ in the scaling of the average shortest-path length with network size: it grows as $\mathcal{O}(\log\log N)$ in USW networks and as $\mathcal{O}(\log N)$ in SW networks. The bipartite networks used to encode node features are generated with $N_f=1000$ features and $\gamma_n=\gamma_f=2.5$, which are the degree-distribution exponents for node and feature vertices, respectively. We set $\langle k_{n}\rangle=10$, the average node degree in the bipartite network.

\paragraph{Real networks.}
For real networks, we consider eight benchmark datasets: Cora, PubMed~\cite{sen2008collective,yang2016revisiting}, Computers, Photo, CS, Physics~\cite{shchur2018pitfalls}, WikiCS~\cite{mernyei2020wiki}, and Flickr~\cite{zeng2020graphsaint}.
Table~\ref{tab:datasets} summarizes the main properties of each dataset. For each network, we infer $\beta$ using Mercator~\cite{garcia2019mercator} and estimate the exponent $\gamma$ of the degree distribution following~\cite{voitalov2019scale}. Based on the classification in~\cite{boguna2020smallworlds}, networks with $\beta<2$ and $2<\gamma<3$ are categorized as ultra-small-world (USW), whereas networks with $\beta<2$ and $\gamma>3$ are categorized as small-world (SW). We used the original dataset features.

\paragraph{GNN training details.} We compare three GNN models: GCN~\cite{kipf2017semisupervised}, GraphSAGE~\cite{hamilton2017}, and GAT~\cite{velickovic2018graph}, each implemented with two layers and 32 hidden dimensions using the PyTorch Geometric library~\cite{FeyLenssen2019}. We train the models using the Adam optimizer with an initial learning rate of 0.01 and a cosine learning-rate scheduler. We also evaluated SGD, but found that it led to weaker performance on some real-world networks. Further details are provided in Appendix~\ref{sec:optimizer}. 

\paragraph{Hyperbolic embeddings of real networks.}  Mercator infers node coordinates by maximizing the likelihood that the observed network was generated by the $\mathbb{S}^1/\mathbb{H}^2$ model. It requires no input parameters beyond the graph itself and returns, for each node, an angular coordinate $\theta_i$ and hidden degree $\kappa_i$, as well as the model parameters $\beta$ and $\mu$. A limitation of Mercator~\cite{garcia2019mercator,jankowski2023d} is the computational cost of embedding large networks, because its time complexity scales as $\mathcal{O}(N^2)$. In this work, we introduce \textit{cuMercator}, an accelerated implementation of Mercator that uses CUDA kernels for GPU execution and negative sampling to approximate the contribution of non-neighbors to the maximum-likelihood calculation. Detailed computational comparisons and estimates of the time required to embed large networks are provided in Appendix~\ref{apx:mercator}.

\subsection{Representational similarity measures}
To investigate why transfer is possible, we compare the similarity between representations learned by GNNs trained on the original network and on its 
GR counterpart. Although many representation-similarity measures are available (see~\cite{kornblith2019similarity}), we restrict our analysis to two measures that capture complementary aspects of representational similarity: Centered Kernel Alignment (CKA)~\cite{cortes2012algorithms,cristianini2001kernel} and Orthogonal Procrustes (OP) analysis~\cite{ding2021grounding,williams2021generalized}. 

For each renormalization level $\ell$, we train a GNN $f_\ell$ on the coarse-grained graph $G_\ell=(V_\ell,E_\ell)$ using labels induced from the original graph $G_0$. To compare this model with a GNN trained directly on $G_0$, we must first put their representations on the same set of nodes. We do this using the hierarchical coarse-graining map $\phi_\ell:V_0\rightarrow V_\ell$, which records which supernode in $G_\ell$ each original node belongs to. Thus, if two original nodes $i$ and $j$ are merged into the same supernode, they are assigned the same coarse-scale representation when we lift back to $V_0$. 
We then compute CKA and OP between the original-scale and lifted coarse-scale representations. CKA compares the node--node geometry induced by the two representations, whereas OP measures how closely they can be aligned by a rotation or reflection. Both measures range from 0 to 1, with larger values indicating greater representational similarity. Formal definitions and further details are provided in Appendix~\ref{app:similarity_measures}.

\subsection{Functional trajectory preservation during training} 

To test whether a renormalized graph preserves the predictive dynamics of training, we compare the class-probability distributions produced during training on the original graph $G_0$ and a coarse graph $G_\ell$. Let $P_{j\to i}(t)$ denote the collection of node-wise categorical distributions over class labels obtained at training step $t$ by a model trained on $G_j$ and evaluated on $G_i$.

First, we compute the \textit{training-graph} divergence
\begin{align}\label{eq:D_JS_TG}
D_\mathrm{JS}^{\mathrm{TG}} = D_{\mathrm{JS}} \left(P_{0\to0}(t), P_{\ell\to0}(t)\right),
\end{align}
where $D_{\mathrm{JS}}$ is the Jensen-Shannon divergence. Small values of $D_\mathrm{JS}^{\mathrm{TG}}$ indicate that the renormalized graph acts as a good training surrogate for the original graph in output space. Second, we examine the \textit{evaluation-graph} divergence: it quantifies how much a model trained on $G_\ell$ changes when evaluated on $G_0$ rather than $G_\ell$,
\begin{align}\label{eq:D_JS_EG}
D_\mathrm{JS}^{\mathrm{EG}} = D_{\mathrm{JS}}\left(U_\ell P_{\ell\to\ell}(t), P_{\ell\to0}(t) \right).
\end{align}
Here, $U_\ell$ denotes the coarse-to-fine lifting operator that maps coarse-node predictions from $G_\ell$ back to the node set of $G_0$. Further definitions and implementation details are provided in Appendix~\ref{apx:trajectories}.

\section{Results}
We report results for synthetic and real networks. In both cases, we first assess whether networks preserve key structural properties along the GR flow. We then evaluate the transfer test accuracy of GNNs trained on renormalized networks. Next, we show that learned representations and the training dynamics remain aligned across graph scales. 

\subsection{Zero-shot GNN transfer on synthetic networks}

Figure~\ref{fig:GR}b presents the topological validation of geometric renormalization for synthetic networks in the small-world regime. The degree distribution, clustering spectrum, and average nearest-neighbor-degree spectrum are preserved during coarse-graining. Global properties, such as the average degree and homophily level, are also maintained. The spectral gap is slightly larger for smaller networks, as expected from its asymptotic vanishing in the large-network limit~\cite{kiwi2018}. Similar results are obtained in the ultra-small-world regime, see Figure~\ref{fig:usw_props} in Appendix~\ref{apx:synth_networks}.

\begin{figure}[h]
	\centering	
    \includegraphics[width=\linewidth]{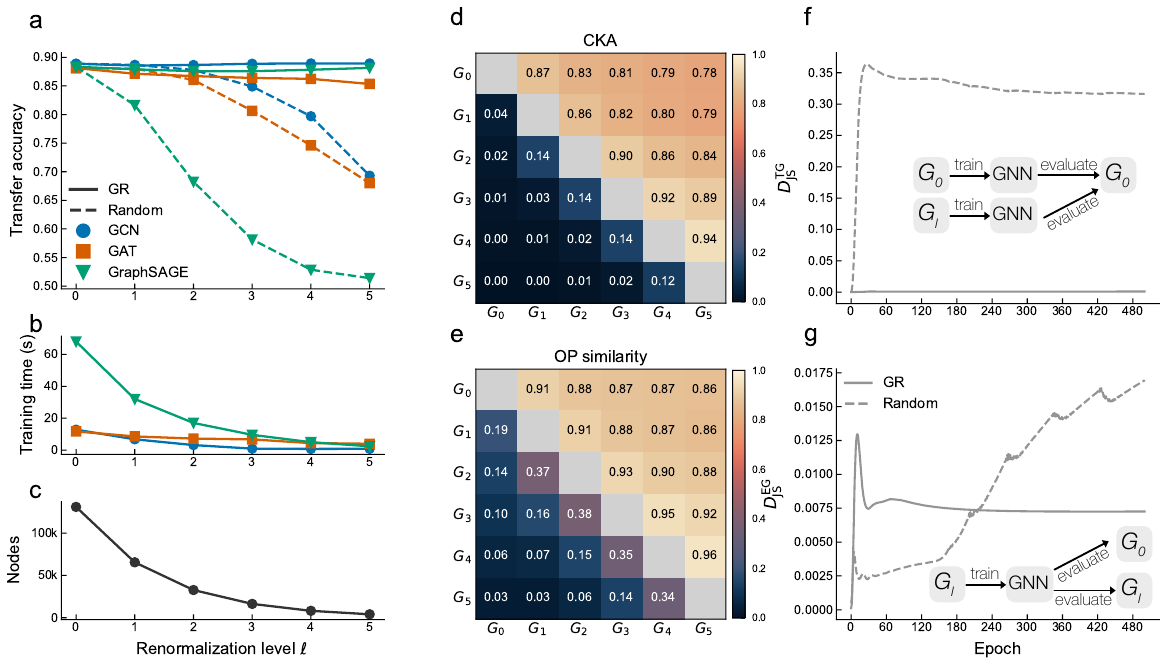}
	\caption{\textbf{Zero-shot GNN transfer on synthetic networks}. (\textbf{a}) Transfer test accuracy as a function of renormalization level $\ell$ for different GNN architectures. Solid lines denote GR and dashed lines denote the random baseline. (\textbf{b}) Training time for GNNs across renormalization levels. (\textbf{c}) Number of nodes at each renormalization level $\ell$. \textbf{(d, e)} Representation similarity between levels, measured by linear CKA and orthogonal Procrustes similarity, respectively (higher is better). The upper-triangular entries correspond to GR and the lower-triangular entries to the random baseline. \textbf{(f, g)} Training alignment of output prediction distributions quantified by Jensen-Shannon divergence $D_\mathrm{JS}$ (lower is better). Panel \textbf{f} shows $D_\mathrm{JS}^{\mathrm{TG}}$ (Eq.~\ref{eq:D_JS_TG}), whereas panel \textbf{g} shows $D_\mathrm{JS}^{\mathrm{EG}}$ (Eq.~\ref{eq:D_JS_EG}). Solid and dashed lines indicate GR and random renormalization, respectively. In panels f and g, we use a GCN model and a GR resolution of $r=32$ to produce the coarse-grained graph $G_\ell$. All results are shown for the synthetic network in the small-world regime.} 
	\label{fig:synthetic_panel}
\end{figure}

In Figure~\ref{fig:synthetic_panel}, we summarize our results for synthetic networks in the small-world regime. Panel~a shows the transfer test accuracy across different renormalization levels. In most cases, it closely matches the test accuracy obtained at the original graph scale. For GCN and GraphSAGE, performance remains high even at a compression factor of $2^5=32$, whereas GAT exhibits a slight decrease. In contrast, the random baseline declines in performance at higher renormalization levels. However, performance does not fall to the level of random chance because the models continue to leverage the original node features, which are not randomized.
Reducing the network size also substantially improves training time, as shown in Fig.~\ref{fig:synthetic_panel}b. For example, for GraphSAGE, training on a coarse-grained network with 4,096 nodes instead of the original network with 131,072 nodes yields a 20-fold speedup, see also Figure~\ref{fig:synthetic_panel}c for the network size at each renormalization level.
Figures~\ref{fig:synthetic_panel}d and~\ref{fig:synthetic_panel}e report the representational-similarity results. GNNs trained independently at different graph scales converge to similar hidden representations, indicating that geometric renormalization preserves the learned representation space across scales. Both linear CKA and OP similarity remain high even under strong compression, whereas random merging produces almost no similarity across scales.
Figures~\ref{fig:synthetic_panel}f,g show the alignment of prediction dynamics across graph scales. The training-graph divergence $D_{\mathrm{JS}}^{\mathrm{TG}}$  (Fig.~\ref{fig:synthetic_panel}f) remains close to zero throughout training for GR, indicating that a model trained on the renormalized graph produces prediction distributions on $G_0$ that closely match those of a model trained directly on $G_0$.  In contrast, the random baseline leads to a rapid and sustained increase in $D_{\mathrm{JS}}^{\mathrm{TG}}$, showing that random coarse graphs do not preserve the output-space training trajectory.
The evaluation-graph divergence $D_{\mathrm{JS}}^{\mathrm{EG}}$ (Fig.~\ref{fig:synthetic_panel}g) remains small for both GR and the random baseline, indicating that models trained on coarse graphs are relatively stable when evaluated across graph resolutions. However, this stability alone is not sufficient to preserve predictive performance: despite similarly low $D_{\mathrm{JS}}^{\mathrm{EG}}$, only GR maintains high transfer test accuracy across renormalization levels, whereas the random baseline leads to a pronounced accuracy degradation (Fig.~\ref{fig:synthetic_panel}a). Together, these results suggest that GR preserves the functional training trajectory.

\subsection{Zero-shot GNN transfer on real networks}
\begin{figure}[h]
	\centering	
    \includegraphics[width=\linewidth]{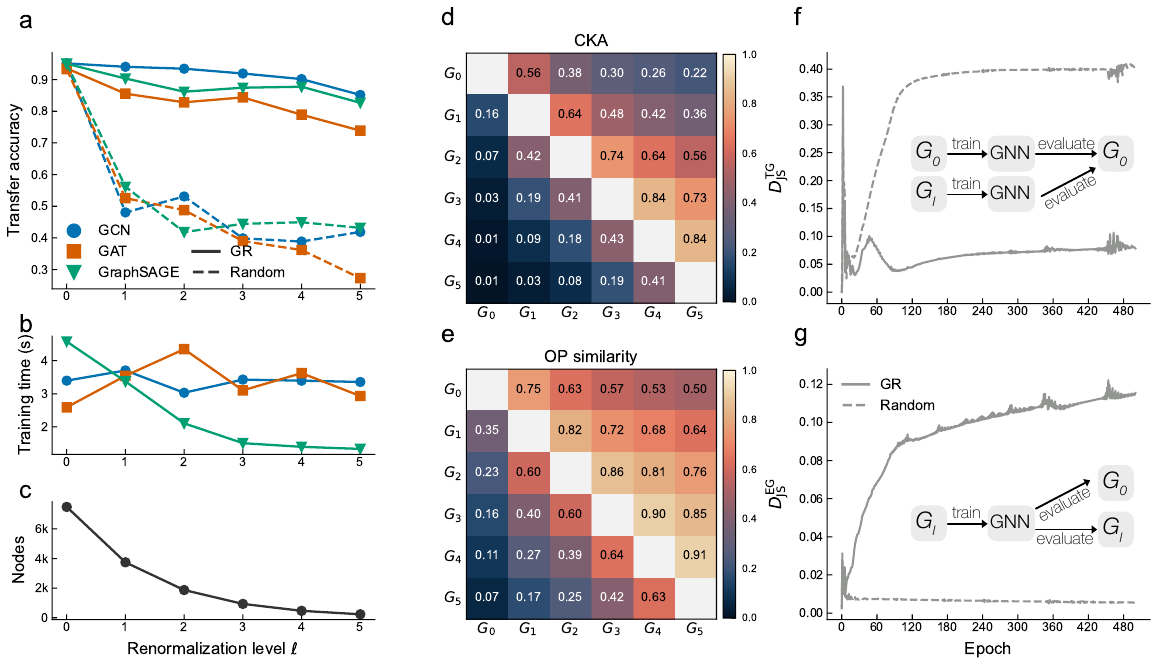}
	\caption{\textbf{Zero-shot GNN transfer on the Photo dataset}. See the caption of Fig.~\ref{fig:synthetic_panel} for more detail.} 
	\label{fig:photo_panel}
\end{figure}

We apply the same methodology to a set of real-world networks. Figure~\ref{fig:photo_panel} summarizes the results for the Photo dataset, while results for the remaining datasets are reported in Appendix~\ref{apx:real_networks}. We also report in Appendix~\ref{apx:realnets_props} the topological validation of the GR transformation.
Overall, we observe qualitatively similar behavior to that seen in the synthetic networks. Under geometric renormalization, transfer test accuracy remains relatively high across renormalization levels (Fig.~\ref{fig:photo_panel}a). Among the architectures considered, the simpler GCN model appears to be the most suitable for weight transfer, outperforming both GAT and GraphSAGE.  Although the transfer accuracy of GraphSAGE decreases by approximately $10\%$, this reduction is accompanied by a roughly fourfold speedup in training time. We note that transfer accuracy on real-world networks is not preserved as strongly as in the synthetic examples, likely because topology, node features, and labels are correlated in a more complex and nontrivial manner~\cite{jankowski2024feature}.

The representation-similarity measures remain higher under GR than under the random baseline (Fig.~\ref{fig:photo_panel}d,e), indicating that the GR graphs better preserve the learned representation structure. As in the synthetic case, representational similarity does not fall to negligible levels under the random baseline because the models continue to leverage the original node features, which are not randomized. The functional trajectories show a qualitatively similar pattern to the synthetic networks (Fig.~\ref{fig:photo_panel}f,g): GR preserves the training-graph trajectory more effectively than the random baseline, while the evaluation-graph divergence remains relatively small. 
In Table~\ref{tab:real_transfer}, we show the transfer test accuracy in the 8 considered datasets.

\begin{table}[t]
\centering
\small
\caption{\textbf{Transfer test accuracy for eight real-world datasets.} For each dataset and level $\ell$, the GCN is trained on the coarse-grained graph $G_\ell$ obtained via GR, and its weights are directly applied, without retraining, to the original graph $G_0$. We report the mean $\pm$ standard deviation of transfer test accuracy over 10 random train/test splits. The bottom row shows the compression ratio $N_0/N_\ell$ at each renormalization level, obtained with a fixed resolution $r=2$ per level.}
\label{tab:real_transfer}
\begin{tabular}{l|c|ccccc}
\toprule
& Original & \multicolumn{5}{c}{Coarse-grained} \\
 & $\ell=0$ & $\ell=1$ & $\ell=2$ & $\ell=3$ & $\ell=4$ & $\ell=5$ \\
\midrule
Computers & 0.92 $\pm$ 0.01 & 0.90 $\pm$ 0.00 & 0.88 $\pm$ 0.01 & 0.87 $\pm$ 0.00 & 0.84 $\pm$ 0.01 & 0.80 $\pm$ 0.03 \\
Cora & 0.87 $\pm$ 0.01 & 0.80 $\pm$ 0.01 & 0.77 $\pm$ 0.01 & 0.77 $\pm$ 0.01 & 0.76 $\pm$ 0.01 & 0.71 $\pm$ 0.02 \\
CS & 0.94 $\pm$ 0.00 & 0.90 $\pm$ 0.00 & 0.87 $\pm$ 0.00 & 0.87 $\pm$ 0.00 & 0.88 $\pm$ 0.01 & 0.88 $\pm$ 0.01 \\
Flickr & 0.52 $\pm$ 0.00 & 0.49 $\pm$ 0.02 & 0.49 $\pm$ 0.01 & 0.50 $\pm$ 0.00 & 0.49 $\pm$ 0.00 & 0.50 $\pm$ 0.00 \\
Photo & 0.95 $\pm$ 0.00 & 0.94 $\pm$ 0.00 & 0.93 $\pm$ 0.01 & 0.92 $\pm$ 0.01 & 0.90 $\pm$ 0.01 & 0.85 $\pm$ 0.02 \\
Physics & 0.96 $\pm$ 0.00 & 0.94 $\pm$ 0.00 & 0.92 $\pm$ 0.00 & 0.92 $\pm$ 0.01 & 0.92 $\pm$ 0.01 & 0.91 $\pm$ 0.01 \\
PubMed & 0.88 $\pm$ 0.00 & 0.85 $\pm$ 0.00 & 0.85 $\pm$ 0.00 & 0.85 $\pm$ 0.01 & 0.84 $\pm$ 0.01 & 0.83 $\pm$ 0.00 \\
WikiCS & 0.83 $\pm$ 0.00 & 0.81 $\pm$ 0.00 & 0.79 $\pm$ 0.01 & 0.75 $\pm$ 0.01 & 0.68 $\pm$ 0.01 & 0.64 $\pm$ 0.01 \\
Compression ratio & $1\times$ & $2\times$ & $4\times$ & $8\times$ & $16\times$ & $32\times$ \\
\bottomrule
\end{tabular}
\end{table}

Several other coarse-graining methods have been proposed in the literature. We compare GR with the Laplacian renormalization group (LRG)~\cite{villegas2023laplacian} and with two edge-contraction pooling methods, MagEdgePool and SpreadEdgePool~\cite{limbeck2026geometry}. 

\begin{wrapfigure}{r}{0.47\textwidth}
    \vspace{-1.0em}
    \centering
    \includegraphics[width=0.47\textwidth]{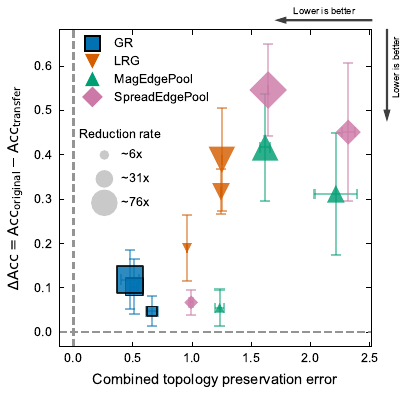}
    \caption{Combined topology preservation error as a function of the difference between the transfer and original test accuracies for different coarse-graining methods for the PubMed dataset. Marker color denotes the coarse-graining method, while marker size represents the compression rate. Larger markers correspond to smaller coarse-grained networks. Results are averaged over 10 train-test splits.} 
    \label{fig:other_coarse_graining}
    \vspace{-3.0em}
\end{wrapfigure}

\newpage 
We repeat the zero-transfer weight evaluation for each method and network reduction rate. Specifically, we measure the difference between the transfer and original test accuracies and the combined topology preservation error. The latter quantifies how well the coarse-graining procedure preserves key topological properties, including the degree distribution, clustering spectrum, average nearest-neighbor degree, average clustering coefficient, and spectral gap. See Appendix~\ref{apx:comparison_coarse_graining} for more details. Figure~\ref{fig:other_coarse_graining} shows that GR produces the smallest accuracy gap, indicating that it maintains high transfer test accuracy even at high compression rates. It also achieves the lowest combined topological error, demonstrating superior preservation of the original network structure. The other methods generally perform worse, except at low compression rates. Together, these results suggest that  GR provides a faithful scaled-down replica of the original graph.

\section{Discussion}
We introduced a train-small, deploy-large protocol for zero-shot GNN transfer through geometric renormalization. Our results show that GNN parameters learned on substantially compressed GR graphs can be transferred directly to the original network while retaining much of the full-scale predictive performance and reducing training cost. This behavior is not explained by graph-size reduction alone: random aggregation substantially degrades transfer, demonstrating that preserving network structure is essential. Models trained across GR scales also learn compatible representation spaces and follow similar predictive trajectories during optimization. These observations indicate that, within a finite range of scales, GR approximately preserves both the representations learned by a GNN and the effective objective it optimizes. 

\section{Limitations}
While the results are promising, we highlight several limitations of our study. First, we renormalize graph structure but not node features. We used simple mean feature aggregation rather than deriving a principled scale transformation for node features, which we leave for future work. Second, we keep the GNN architecture and hyperparameters fixed across scales; whether hyperparameters optimized at a lower scale remain optimal at the original scale is therefore unknown. Third, transfer may depend on the dimensionality of the underlying hyperbolic representation~\cite{FerraMarcus2026,almagro2022detecting,jankowski2023d}. However, geometric renormalization is currently limited to a one-dimensional similarity space. Fourth, we focus on homophilic graphs, whether a similar scale transfer approach is possible in heterophilic settings remains open. Finally, we study node-level learning tasks; extending the framework to graph classification~\cite{koke2026graph}, link prediction, and other graph-learning problems is an important direction for future work.

\section*{Acknowledgments}
The authors acknowledge support from Grants PID2022-137505NB-C22, PID2024-156576OB-C33, and PID2023-147198NB-I00, funded by MICIU/AEI/10.13039/501100011033 and by ERDF/EU. M.~B. acknowledges support from the Agency for Management of University and Research Grants of the Generalitat de Catalunya through the Acad\`emia d’Excel·l\`encia grant. M.~W. acknowledges partial support from NSF award DMS-2406905, an Alfred P. Sloan Research Fellowship in Mathematics, the AI2050 program at Schmidt Sciences (Grant G-25-69786), and an Aramont Fellowship for Emerging Science Research.

\bibliographystyle{unsrtnat}
\bibliography{reference}

@book{serrano2022shortest,
  title={The Shortest Path to Network Geometry: A Practical Guide to Basic Models and Applications},
  author={Serrano, M {\'A}ngeles and Bogu{\~n}{\'a}, Mari{\'a}n},
  year={2022},
  publisher={Cambridge University Press}
}

@inproceedings{
kipf2017semisupervised,
title={Semi-Supervised Classification with Graph Convolutional Networks},
author={Thomas N. Kipf and Max Welling},
booktitle={International Conference on Learning Representations},
year={2017},
url={https://openreview.net/forum?id=SJU4ayYgl}
}

@inproceedings{huang2021scaling,
  title     = {Scaling Up Graph Neural Networks Via Graph Coarsening},
  author    = {Huang, Zengfeng and Zhang, Shengzhong and Xi, Chong and Liu, Tang and Zhou, Min},
  booktitle = {Proceedings of the 27th ACM SIGKDD Conference on Knowledge Discovery and Data Mining (KDD '21)},
  pages     = {675--684},
  year      = {2021},
  doi       = {10.1145/3447548.3467256}
}

@article{shchur2018pitfalls,
  title   = {Pitfalls of Graph Neural Network Evaluation},
  author  = {Shchur, Oleksandr and Mumme, Max and Bojchevski, Aleksandar and G{\"u}nnemann, Stephan},
  journal = {arXiv preprint arXiv:1811.05868},
  year    = {2018},
  doi     = {10.48550/arXiv.1811.05868}
}

@article{joly2024graph,
  title={Graph coarsening with message-passing guarantees},
  author={Joly, Antonin and Keriven, Nicolas},
  journal={Advances in Neural Information Processing Systems},
  volume={37},
  pages={114902--114927},
  year={2024}
}

@article{ruiz2020graphon,
  title={Graphon neural networks and the transferability of graph neural networks},
  author={Ruiz, Luana and Chamon, Luiz and Ribeiro, Alejandro},
  journal={Advances in Neural Information Processing Systems},
  volume={33},
  pages={1702--1712},
  year={2020}
}

@article{koke2026graph,
  title={Graph Neural Networks Are Not Continuous Across Graph Resolutions},
  author={Koke, Christian and Shen, Yuesong and Saroha, Abhishek and Eisenberger, Marvin and Rieck, Bastian and Bronstein, Michael and Cremers, Daniel},
  journal={arXiv preprint arXiv:2605.31315},
  year={2026}
}

@inproceedings{yehudai2021local,
  title={From local structures to size generalization in graph neural networks},
  author={Yehudai, Gilad and Fetaya, Ethan and Meirom, Eli and Chechik, Gal and Maron, Haggai},
  booktitle={International Conference on Machine Learning},
  pages={11975--11986},
  year={2021},
  organization={PMLR}
}

@inproceedings{hamilton2017,
 author = {Hamilton, Will and Ying, Zhitao and Leskovec, Jure},
 booktitle = {Advances in Neural Information Processing Systems},
 editor = {I. Guyon and U. Von Luxburg and S. Bengio and H. Wallach and R. Fergus and S. Vishwanathan and R. Garnett},
 pages = {},
 publisher = {Curran Associates, Inc.},
 title = {Inductive Representation Learning on Large Graphs},
 url = {https://proceedings.neurips.cc/paper_files/paper/2017/file/5dd9db5e033da9c6fb5ba83c7a7ebea9-Paper.pdf},
 volume = {30},
 year = {2017}
}

@article{loukas2019graph,
  title={Graph reduction with spectral and cut guarantees},
  author={Loukas, Andreas},
  journal={Journal of Machine Learning Research},
  volume={20},
  number={116},
  pages={1--42},
  year={2019}
}

@article{chen2022graph,
  title={Graph coarsening: from scientific computing to machine learning},
  author={Chen, Jie and Saad, Yousef and Zhang, Zechen},
  journal={SeMA Journal},
  volume={79},
  number={1},
  pages={187--223},
  year={2022},
  publisher={Springer}
}

@article{boguna2021network,
  title={Network geometry},
  author={Boguna, Marian and Bonamassa, Ivan and De Domenico, Manlio and Havlin, Shlomo and Krioukov, Dmitri and Serrano, M {\'A}ngeles},
  journal={Nature Reviews Physics},
  volume={3},
  number={2},
  pages={114--135},
  year={2021},
  publisher={Nature Publishing Group UK London}
}

@article{serrano2008self,
  title={Self-similarity of complex networks and hidden metric spaces},
  author={Serrano, M {\'A}ngeles and Krioukov, Dmitri and Bogun{\'a}, Mari{\'a}n},
  journal={Physical Review Letters},
  volume={100},
  number={7},
  pages={078701},
  year={2008},
  publisher={APS}
}

@article{krioukov2010hyperbolic,
  title={Hyperbolic geometry of complex networks},
  author={Krioukov, Dmitri and Papadopoulos, Fragkiskos and Kitsak, Maksim and Vahdat, Amin and Bogun{\'a}, Mari{\'a}n},
  journal={Physical Review E—Statistical, Nonlinear, and Soft Matter Physics},
  volume={82},
  number={3},
  pages={036106},
  year={2010},
  publisher={APS}
}

@article{garcia2018multiscale,
  title={Multiscale unfolding of real networks by geometric renormalization},
  author={Garc{\'\i}a-P{\'e}rez, Guillermo and Bogu{\~n}{\'a}, Mari{\'a}n and Serrano, M {\'A}ngeles},
  journal={Nature Physics},
  volume={14},
  number={6},
  pages={583--589},
  year={2018},
  publisher={Nature Publishing Group UK London}
}

@ARTICLE{Aliakbarisani2026,
  author={Aliakbarisani, Roya and Jankowski, Robert and Serrano, M. Ángeles and Boguñá, Marián},
  journal={IEEE Transactions on Neural Networks and Learning Systems}, 
  title={HypBench: Hyperbolic Benchmark for Graph Neural Network Performance}, 
  year={2026},
  volume={},
  number={},
  pages={1-14},
  doi={10.1109/TNNLS.2026.3697597}}

@inproceedings{kornblith2019similarity,
  title={Similarity of neural network representations revisited},
  author={Kornblith, Simon and Norouzi, Mohammad and Lee, Honglak and Hinton, Geoffrey},
  booktitle={International conference on machine learning},
  pages={3519--3529},
  year={2019},
  organization={PMLR}
}

@article{ruiz2023transferability,
  title={Transferability properties of graph neural networks},
  author={Ruiz, Luana and Chamon, Luiz FO and Ribeiro, Alejandro},
  journal={IEEE Transactions on Signal Processing},
  volume={71},
  pages={3474--3489},
  year={2023},
  publisher={IEEE}
}

@article{ying2018hierarchical,
  title={Hierarchical graph representation learning with differentiable pooling},
  author={Ying, Zhitao and You, Jiaxuan and Morris, Christopher and Ren, Xiang and Hamilton, Will and Leskovec, Jure},
  journal={Advances in neural information processing systems},
  volume={31},
  year={2018}
}

@InProceedings{gao2019graph,
  title = 	 {Graph U-Nets},
  author =       {Gao, Hongyang and Ji, Shuiwang},
  booktitle = 	 {Proceedings of the 36th International Conference on Machine Learning},
  pages = 	 {2083--2092},
  year = 	 {2019},
  editor = 	 {Chaudhuri, Kamalika and Salakhutdinov, Ruslan},
  volume = 	 {97},
  series = 	 {Proceedings of Machine Learning Research},
  month = 	 {09--15 Jun},
  publisher =    {PMLR},
  url = 	 {https://proceedings.mlr.press/v97/gao19a.html}
}

@article{gabrielli2025network,
  title={Network renormalization},
  author={Gabrielli, Andrea and Garlaschelli, Diego and Patil, Subodh P and Serrano, M {\'A}ngeles},
  journal={Nature Reviews Physics},
  volume={7},
  number={4},
  pages={203--219},
  year={2025},
  publisher={Nature Publishing Group UK London}
}

@article{song2005self,
  title={Self-similarity of complex networks},
  author={Song, Chaoming and Havlin, Shlomo and Makse, Hernan A},
  journal={Nature},
  volume={433},
  number={7024},
  pages={392--395},
  year={2005},
  publisher={Nature Publishing Group UK London}
}

@article{zheng2020geometric,
  title={Geometric renormalization unravels self-similarity of the multiscale human connectome},
  author={Zheng, Muhua and Allard, Antoine and Hagmann, Patric and Alem{\'a}n-G{\'o}mez, Yasser and Serrano, M {\'A}ngeles},
  journal={Proceedings of the National Academy of Sciences},
  volume={117},
  number={33},
  pages={20244--20253},
  year={2020},
  publisher={National Academy of Sciences}
}

@article{zheng2024geometric,
  title={Geometric renormalization of weighted networks},
  author={Zheng, Muhua and Garc{\'\i}a-P{\'e}rez, Guillermo and Bogu{\~n}{\'a}, Mari{\'a}n and Serrano, M {\'A}ngeles},
  journal={Communications Physics},
  volume={7},
  number={1},
  pages={97},
  year={2024},
  publisher={Nature Publishing Group UK London}
}

@article{villegas2023laplacian,
  title={Laplacian renormalization group for heterogeneous networks},
  author={Villegas, Pablo and Gili, Tommaso and Caldarelli, Guido and Gabrielli, Andrea},
  journal={Nature Physics},
  volume={19},
  number={3},
  pages={445--450},
  year={2023},
  publisher={Nature Publishing Group UK London}
}

@article{nurisso2025higher,
  title={Higher-order Laplacian renormalization},
  author={Nurisso, Marco and Morandini, Marta and Lucas, Maxime and Vaccarino, Francesco and Gili, Tommaso and Petri, Giovanni},
  journal={Nature Physics},
  volume={21},
  number={4},
  pages={661--668},
  year={2025},
  publisher={Nature Publishing Group UK London}
}

@article{garuccio2023multiscale,
  title={Multiscale network renormalization: Scale-invariance without geometry},
  author={Garuccio, Elena and Lalli, Margherita and Garlaschelli, Diego},
  journal={Physical Review Research},
  volume={5},
  number={4},
  pages={043101},
  year={2023},
  publisher={APS}
}

@article{cortes2012algorithms,
  title={Algorithms for learning kernels based on centered alignment},
  author={Cortes, Corinna and Mohri, Mehryar and Rostamizadeh, Afshin},
  journal={The Journal of Machine Learning Research},
  volume={13},
  pages={795--828},
  year={2012},
  publisher={JMLR. org}
}

@article{cristianini2001kernel,
  title={On kernel-target alignment},
  author={Cristianini, Nello and Shawe-Taylor, John and Elisseeff, Andre and Kandola, Jaz},
  journal={Advances in neural information processing systems},
  volume={14},
  year={2001}
}

@article{ding2021grounding,
  title={Grounding representation similarity through statistical testing},
  author={Ding, Frances and Denain, Jean-Stanislas and Steinhardt, Jacob},
  journal={Advances in Neural Information Processing Systems},
  volume={34},
  pages={1556--1568},
  year={2021}
}

@article{williams2021generalized,
  title={Generalized shape metrics on neural representations},
  author={Williams, Alex H and Kunz, Erin and Kornblith, Simon and Linderman, Scott},
  journal={Advances in neural information processing systems},
  volume={34},
  pages={4738--4750},
  year={2021}
}

@article{jankowski2023d,
  title={The D-Mercator method for the multidimensional hyperbolic embedding of real networks},
  author={Jankowski, Robert and Allard, Antoine and Boguna, Marian and Serrano, M {\'A}ngeles},
  journal={Nature Communications},
  volume={14},
  number={1},
  pages={7585},
  year={2023},
  publisher={Nature Publishing Group UK London}
}

@article{garcia2019mercator,
  title={Mercator: uncovering faithful hyperbolic embeddings of complex networks},
  author={Garc{\'\i}a-P{\'e}rez, Guillermo and Allard, Antoine and Serrano, M {\'A}ngeles and Bogu{\~n}{\'a}, Mari{\'a}n},
  journal={New Journal of Physics},
  volume={21},
  number={12},
  pages={123033},
  year={2019},
  publisher={IOP Publishing}
}

@article{muscoloni2017machine,
  title={Machine learning meets complex networks via coalescent embedding in the hyperbolic space},
  author={Muscoloni, Alessandro and Thomas, Josephine Maria and Ciucci, Sara and Bianconi, Ginestra and Cannistraci, Carlo Vittorio},
  journal={Nature Communications},
  volume={8},
  number={1},
  pages={1615},
  year={2017},
  publisher={Nature Publishing Group UK London}
}

@inproceedings{
velickovic2018graph,
title={Graph Attention Networks},
author={Petar Veličković and Guillem Cucurull and Arantxa Casanova and Adriana Romero and Pietro Liò and Yoshua Bengio},
booktitle={International Conference on Learning Representations},
year={2018},
url={https://openreview.net/forum?id=rJXMpikCZ},
}

@article{boguna2020smallworlds,
  title     = {Small worlds and clustering in spatial networks},
  author    = {Bogu{\~n}{\'a}, Marian and Krioukov, Dmitri and Almagro, Pedro and Serrano, M. {\'A}ngeles},
  journal   = {Physical Review Research},
  volume    = {2},
  number    = {2},
  pages     = {023040},
  year      = {2020},
  month     = {Apr},
  publisher = {American Physical Society},
  doi       = {10.1103/PhysRevResearch.2.023040}
}

@article{limbeck2026geometry,
  title={Geometry-aware edge pooling for graph neural networks},
  author={Limbeck, Katharina and Mezrag, Lydia and Wolf, Guy and Rieck, Bastian},
  journal={Advances in Neural Information Processing Systems},
  volume={38},
  pages={157770--157808},
  year={2026}
}

@article{voitalov2019scale,
  title={Scale-free networks well done},
  author={Voitalov, Ivan and Van Der Hoorn, Pim and Van Der Hofstad, Remco and Krioukov, Dmitri},
  journal={Physical Review Research},
  volume={1},
  number={3},
  pages={033034},
  year={2019},
  publisher={APS}
}

@inproceedings{FeyLenssen2019,
  title={Fast Graph Representation Learning with {PyTorch Geometric}},
  author={Fey, Matthias and Lenssen, Jan E.},
  booktitle={ICLR Workshop on Representation Learning on Graphs and Manifolds},
  year={2019},
}

@Article{FerraMarcus2026,
author={Ferr{\`a} Marc{\'u}s, Aina
and Jankowski, Robert
and Vila-Mi{\~{n}}ana, Meritxell
and Casacuberta, Carles
and Serrano, M. {\'A}ngeles},
title={Chordless cycle filtrations for dimensionality detection in complex networks via topological data analysis},
journal={Nature Communications},
year={2026},
volume={17},
number={1},
pages={6105},
}

@article{almagro2022detecting,
  title={Detecting the ultra low dimensionality of real networks},
  author={Almagro, Pedro and Bogu{\~n}{\'a}, Mari{\'a}n and Serrano, M {\'A}ngeles},
  journal={Nature Communications},
  volume={13},
  number={1},
  pages={6096},
  year={2022},
  publisher={Nature Publishing Group UK London}
}

@article{sen2008collective,
  title={Collective classification in network data},
  author={Sen, Prithviraj and Namata, Galileo and Bilgic, Mustafa and Getoor, Lise and Galligher, Brian and Eliassi-Rad, Tina},
  journal={AI Magazine},
  volume={29},
  number={3},
  pages={93--106},
  year={2008}
}

@inproceedings{yang2016revisiting,
  title={Revisiting semi-supervised learning with graph embeddings},
  author={Yang, Zhilin and Cohen, William and Salakhudinov, Ruslan},
  booktitle={International conference on machine learning},
  pages={40--48},
  year={2016},
  organization={PMLR}
}

@article{mernyei2020wiki,
  title={Wiki-CS: A Wikipedia-based benchmark for graph neural networks},
  author={Mernyei, P{\'e}ter and Cangea, C{\u{a}}t{\u{a}}lina},
  journal={arXiv preprint arXiv:2007.02901},
  year={2020}
}

@inproceedings{zeng2020graphsaint,
  title={{GraphSAINT}: Graph sampling based inductive learning method},
  author={Zeng, Hanqing and Zhou, Hongkuan and Srivastava, Ajitesh and Kannan, Rajgopal and Prasanna, Viktor},
  booktitle={International Conference on Learning Representations (ICLR)},
  year={2020}
}

@article{kiwi2018,
author = {Marcos Kiwi and Dieter Mitsche},
title = {{Spectral gap of random hyperbolic graphs and related parameters}},
volume = {28},
journal = {The Annals of Applied Probability},
number = {2},
publisher = {Institute of Mathematical Statistics},
pages = {941 -- 989},
year = {2018},
doi = {10.1214/17-AAP1323},
URL = {https://doi.org/10.1214/17-AAP1323}
}

@article{newman2003structure,
  title={The structure and function of complex networks},
  author={Newman, Mark EJ},
  journal={SIAM Review},
  volume={45},
  number={2},
  pages={167--256},
  year={2003},
  publisher={SIAM}
}

@article{grattarola2022understanding,
  title={Understanding pooling in graph neural networks},
  author={Grattarola, Daniele and Zambon, Daniele and Bianchi, Filippo Maria and Alippi, Cesare},
  journal={IEEE transactions on neural networks and learning systems},
  volume={35},
  number={2},
  pages={2708--2718},
  year={2022},
  publisher={IEEE}
}

@article{fenggraph,
  title={Graph Pooling via Ricci Flow},
  author={Feng, Amy and Weber, Melanie},
  journal={Transactions on Machine Learning Research},
  year={2024}
}

@article{li2023metadata,
  title={A metadata-driven approach to understand graph neural networks},
  author={Li, Ting Wei and Mei, Qiaozhu and Ma, Jiaqi},
  journal={Advances in Neural Information Processing Systems},
  volume={36},
  pages={15320--15340},
  year={2023}
}

@article{fesser2025performance,
  title={Performance Heterogeneity in Graph Neural Networks: Lessons for Architecture Design and Preprocessing},
  author={Fesser, Lukas and Weber, Melanie},
  journal={arXiv preprint arXiv:2503.00547},
  year={2025}
}

@article{jankowski2024feature,
  title={Feature-aware ultra-low dimensional reduction of real networks},
  author={Jankowski, Robert and Hozhabrierdi, Pegah and Bogu{\~n}{\'a}, Mari{\'a}n and Serrano, M {\'A}ngeles},
  journal={npj Complexity},
  volume={1},
  number={1},
  pages={13},
  year={2024},
  publisher={Nature Publishing Group UK London}
}

\newpage
\appendix
\section{Appendix}

\subsection{Geometric renormalization in detail}\label{apx:GR}
Here, we describe the renormalizability of the $\mathbb{S}^1$ model. For more details, such as the semigroup structure of the coarse-graining step and the solution in the power-law approximation, please refer to the Supplementary Information of~\cite{garcia2018multiscale}. The renormalized networks remain maximally congruent with the
hidden metric space model by assigning a new hidden degree $\kappa_i^{(l+1)}$ to supernode $i$ in layer $l+1$ as a function of the hidden degrees of nodes it contains in layer $l$ according to
\begin{align}
    \kappa_i^{(l+1)} = \left(\sum\limits_{j=1}^r \left(\kappa_j^{(l)}\right)^\beta\right)^{1/\beta}
\end{align}
as well as an angular coordinate $\theta_i^{(l+1)}$ given by
\begin{align}
    \theta_i^{(l+1)} = \left(\frac{\sum\limits_{j=1}^r \left(\theta_j^{(l)} \kappa_j^{(l)}\right)^\beta}{\sum\limits_{j=1}^r \left( \kappa_j^{(l)}\right)^\beta}\right)^{1/\beta}.
\end{align}
The global parameters also need to be rescaled as $\mu^{(l+1)} = \mu^{(l)}/r$, $\beta^{(l+1)} = \beta^{(l)}$, and $R^{(l+1)} = R^{(l)}/r$. This implies that the probability $p_{ij}^{(l+1)}$ for two supernodes $i$ and $j$ to be connected in layer $l+1$ maintains its original form (Eq.~\ref{eq:s1}). This applies both to the
model and to real networks as long as they admit a good embedding. 
During the renormalization flow, the average degree increases according to $\langle k \rangle^{(l+1)} = r^\nu \langle k \rangle^{(l)}$, where $\nu$ is a scaling exponent determined by the connectivity structure of the original network. We therefore adjust the average degree of the downscaled network replicas. Specifically, to match the average degree of a renormalized network to that of the original network, we prune links according to the underlying metric model with which networks at all layers are congruent.

\subsection{Representational similarity measures}\label{app:similarity_measures}

Let $n=|V_0|$ and let $d$ denote the common representation dimension of all models, which use the same GNN architecture. We write
\begin{align}
Z_0\in\mathbb{R}^{n\times d},
\qquad
Z_\ell\in\mathbb{R}^{|V_\ell|\times d}
\end{align}
for the node representations learned on the original and coarse-grained graphs. The coarse-scale representations are lifted to the original node set 
so that $Z_\ell^\uparrow\in\mathbb{R}^{n\times d}$. We center each feature across nodes, obtaining $X_0$ and $X_\ell^\uparrow$.

We compute linear CKA as
\begin{align}
    \mathrm{CKA}(X_0,X_\ell^\uparrow)
    =
    \frac{\|X_0^\top X_\ell^\uparrow\|_F^2}
    {\|X_0^\top X_0 |_F \, \|(X_\ell^\uparrow)^\top X_\ell^\uparrow\|_F},
\end{align}
and normalized Orthogonal Procrustes similarity as
\begin{align}
    \mathrm{OP}(X_0,X_\ell^\uparrow) = \frac{\|X_0^\top X_\ell^\uparrow\|_*}{\|X_0\|_F\|X_\ell^\uparrow\|_F},
\end{align}
Here, $\|\cdot\|_F$ is the Frobenius norm, i.e., the square root of the sum of squared matrix entries, and $\|\cdot\|_*$ is the nuclear norm, i.e., the sum of the singular values.

For nonzero representation matrices, both measures are in $[0,1]$. A value of 1 indicates maximal similarity under the corresponding criterion, whereas 0 indicates no shared linear structure. CKA compares the geometry induced among nodes and is invariant to isotropic scaling and orthogonal transformations. OP measures the maximal similarity after rotating or reflecting one representation.

\subsection{Functional trajectory diagnostics}\label{apx:trajectories}

Representation-similarity measures characterize learned embeddings but do not directly determine whether training on a graph-renormalization replica follows the same predictive trajectory as training on the original graph. We therefore compare model outputs throughout training rather than gradients in parameter space, which may be sensitive to parameterization and training noise.

Let $Y_v\in\{1,\ldots,C\}$ denote the class-label random variable for node $v$, where $C$ is the number of classes. Let $\omega_t^{(j)}$ denote the parameters of a GNN after training for $t$ steps on graph $G_j$. For a node $v$ in the evaluation graph $G_i$, define
\begin{align}
P_{j\to i}^{(v)}(t)
=
\operatorname{softmax}\!\left(
f_{\omega_t^{(j)}}(G_i)_v
\right).
\end{align}
We use $P_{j\to i}(t)$ to denote the collection of these node-level distributions over the evaluated node set. In particular, $P_{0\to0}(t)$ corresponds to training and evaluation on $G_0$, $P_{\ell\to\ell}(t)$ to training and evaluation on $G_\ell$, and $P_{\ell\to0}(t)$ to training on $G_\ell$ followed by evaluation on $G_0$.
\begin{align}
D_{\mathrm{JS}}^{\mathrm{TG}}(t)
=
D_{\mathrm{JS}}\left(P_{0\to0}(t),P_{\ell\to0}(t)\right).
\end{align}
This quantity isolates the effect of replacing $G_0$ with $G_\ell$ during training while keeping the evaluation graph fixed. Small values indicate that $G_\ell$ acts as an effective training surrogate for $G_0$ in output space.
\begin{align}
D_{\mathrm{JS}}^{\mathrm{EG}}(t)
&=
D_{\mathrm{JS}}\left(
U_\ell P_{\ell\to\ell}(t),
P_{\ell\to0}(t)
\right).
\end{align}
This quantity keeps the trained model fixed and compares its predictions at the coarse and original graph resolutions. The operator $U_\ell$ maps each coarse-node prediction to the corresponding nodes of $G_0$. Small values therefore indicate that the coarse-trained model transfers stably to the original graph.
Both divergences are evaluated throughout training. Together, they distinguish the quality of $G_\ell$ as a training surrogate from the stability of evaluating a coarse-trained model at the original graph resolution.

\subsection{Properties of real networks}\label{apx:real_nets}

\begin{table}[ht]
\centering
\small
\caption{Properties of the real datasets. $N$: number of nodes; $E$: number of edges; $N_f$: number of node features; $N_c$: number of classes; $\bar{k}$: average degree; $\mathcal{H}$: homophily ratio; $\gamma$: degree-distribution exponent; $\beta$: clustering-coefficient parameter of the $\mathbb{S}^1/\mathbb{H}^2$ model, inferred with Mercator; $\bar{c}$: average clustering coefficient; Regime: structural classification of the network as ultra-small-world (USW) or small-world (SW), following~\cite{boguna2020smallworlds}, based on the inferred values of $\beta$ and $\gamma$.} 
\label{tab:datasets}
\begin{tabular}{lrrrrrrrrrc}
\toprule
Dataset & \multicolumn{1}{c}{$N$} & \multicolumn{1}{c}{$E$} & \multicolumn{1}{c}{$N_f$} & \multicolumn{1}{c}{$N_c$} & \multicolumn{1}{c}{$\bar{k}$} & \multicolumn{1}{c}{$\mathcal{H}$} & \multicolumn{1}{c}{$\gamma$} & \multicolumn{1}{c}{$\beta$} & \multicolumn{1}{c}{$\bar{c}$} & \multicolumn{1}{c}{Regime} \\
\midrule
Computers & 13,381 & 245,778 & 767 & 10 & 36.7 & 0.78 & 2.96 & 1.46 & 0.36 & USW \\
Cora & 2,485 & 5,069 & 1433 & 7 & 4.1 & 0.81 & 3.08 & 1.54 & 0.29 & SW \\
CS & 18,333 & 81,894 & 6805 & 15 & 8.9 & 0.81 & 6.11 & 1.72 & 0.37 & SW \\
Flickr & 89,250 & 449,878 & 500 & 7 & 10.1 & 0.32 & 2.64 & 1.01 & 0.08 & USW \\
Photo & 7,487 & 119,043 & 745 & 8 & 31.8 & 0.83 & 2.99 & 1.65 & 0.41 & USW \\
Physics & 34,493 & 247,962 & 8415 & 5 & 14.4 & 0.93 & 5.0 & 1.71 & 0.38 & SW \\
PubMed & 19,717 & 44,324 & 500 & 3 & 4.5 & 0.8 & 4.18 & 1.12 & 0.11 & SW \\
WikiCS & 11,311 & 215,554 & 300 & 10 & 38.1 & 0.65 & 3.72 & 1.61 & 0.5 & SW \\
\bottomrule
\end{tabular}
\end{table}

\subsection{Comparison with other coarse-graining methods}\label{apx:comparison_coarse_graining}
We compare geometric renormalization (GR) with the Laplacian renormalization group (LRG), which identifies appropriate spatiotemporal scales in heterogeneous networks~\cite{villegas2023laplacian}, and with two edge-contraction pooling methods, MagEdgePool and SpreadEdgePool. The latter methods iteratively contract the least important edges and are computed before GNN training~\cite{limbeck2026geometry}.

For LRG, we first fix the network's temporal resolution parameter $\tau$. For GR, we choose the resolution parameter $r$ so that the final size of the coarse-grained network matches the target compression level. We use the corresponding pooling ratio for MagEdgePool and SpreadEdgePool.

\begin{figure}[h]
	\centering
	\includegraphics[width=\linewidth]{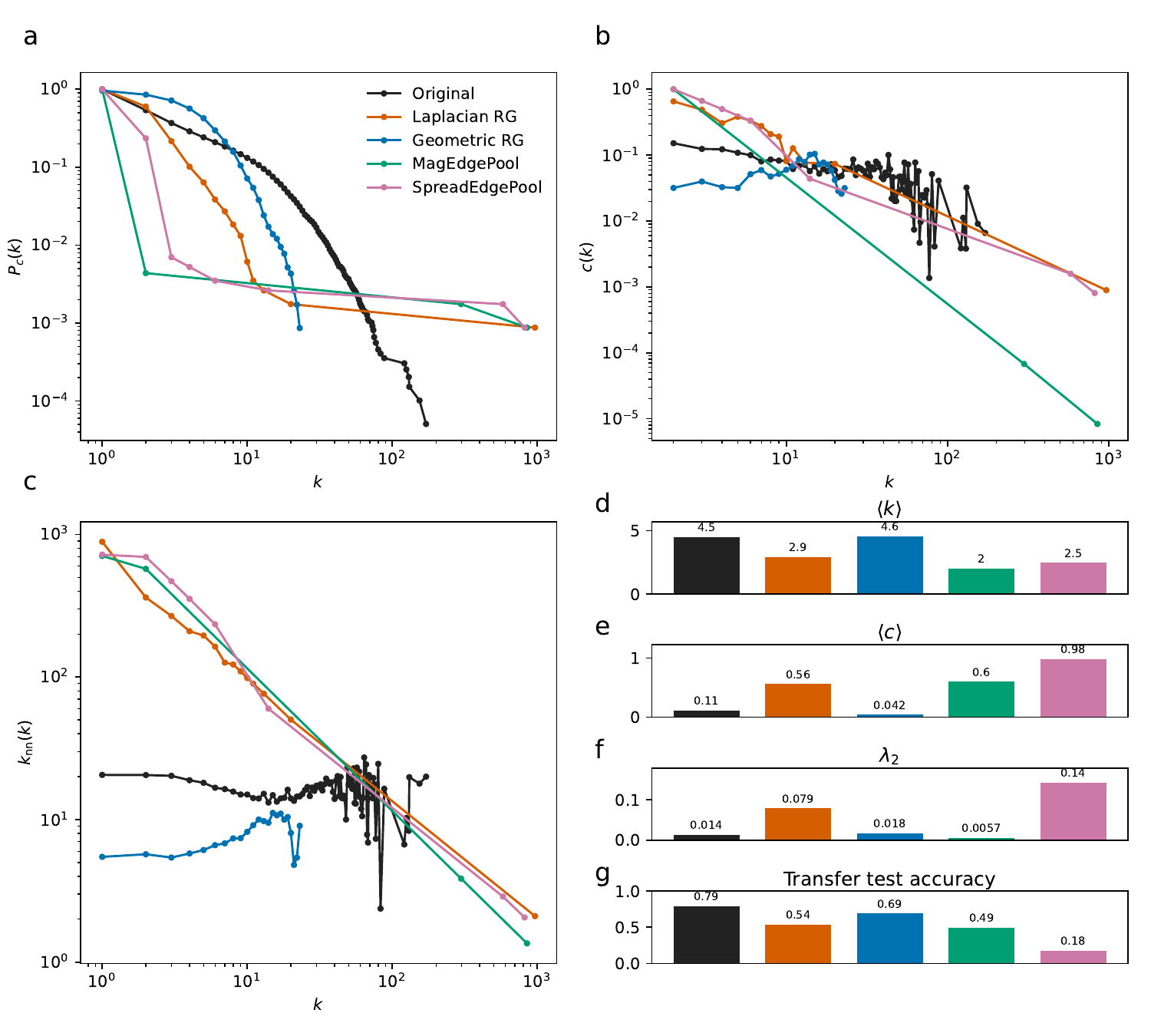}
    \caption{Comparison of coarse-graining methods on the PubMed dataset. We report: (a) the complementary cumulative degree distribution, (b) the clustering spectrum, (c) the average-neighbor-degree spectrum, (d) the average degree, (e) the average clustering coefficient, (f) the spectral gap, and (g) the transfer test accuracy. For LRG, we use $\tau=1.2$, which yields approximately 1100 nodes, compressing the original network by a factor of approximately 17. For GR, MagEdgePool, and SpreadEdgePool, we match this number of nodes by varying the resolution parameter $r$ for GR and the pooling ratio for MagEdgePool and SpreadEdgePool. For the transfer task, we train a two-layer GCN with 32 hidden dimensions.}
	\label{fig:pubmed_comparison}
\end{figure}

We first compare the topological properties of the coarse-grained networks produced by each method. In Figure~\ref{fig:pubmed_comparison}a-f, we show the degree distribution, clustering spectrum, average nearest-neighbor degree spectrum, and global network properties, including the average degree, average clustering coefficient, and spectral gap. Overall, GR preserves both local and global network properties more accurately than the other methods.
We then evaluate transfer accuracy (Fig.~\ref{fig:pubmed_comparison}g). GR yields the smallest drop in transfer accuracy compared with a GNN trained on the original PubMed network.

We extend the comparison by varying the parameter $\tau$, which controls the compression rate. In Figure~\ref{fig:other_coarse_graining}, we consider $\tau \in \{0.5, 1.6, 2\}$, corresponding to approximate reductions in network size of $6\times$, $31\times$, and $76\times$, respectively. We compute the combined topology-preservation error as the sum of the normalized discrepancies across all topology metrics except average degree, because GR explicitly preserves the average degree through network pruning~\cite{garcia2018multiscale}. This combined topology preservation error is defined as the sum of these normalized topology errors as 
\begin{align}
\mathrm{Err}_{\mathrm{topo}}(G_\ell, G_0)
= \widetilde{\mathrm{Err}}_{P(k)} + \widetilde{\mathrm{Err}}_{c(k)} + \widetilde{\mathrm{Err}}_{k_{\mathrm{nn}}(k)} + \widetilde{\mathrm{Err}}_{\bar{c}} + \widetilde{\mathrm{Err}}_{\lambda_2},
\end{align}
where $\widetilde{\mathrm{Err}}_{i}$ is the normalized error for topology statistic $i$, obtained by dividing the corresponding raw error by the maximum raw error of the same statistic across all coarse-grained graphs included in the comparison. Lower values indicate better preservation of the original topology. The degree-distribution term $\mathrm{Err}_{P(k)}$ is the Anderson--Darling statistic between the node-degree samples of $G_0$ and $G_\ell$. The spectrum terms $\mathrm{Err}_{c(k)}$ and $\mathrm{Err}_{k_{\mathrm{nn}}(k)}$ are root-mean-square errors (RMSE) between the original and coarse-grained degree-dependent clustering spectrum $c(k)$ and average-neighbor-degree spectrum $k_{\mathrm{nn}}(k)$, respectively. The mean-clustering term is the relative error $\mathrm{Err}_{\bar{c}}
=\frac{\left|\bar{c}_\ell - \bar{c}_0\right|}{\bar{c}_0}$, and the spectral-gap term is the absolute log-ratio error
$\mathrm{Err}_{\lambda_2}=\left|\log \frac{\lambda_{2,\ell}+\epsilon}{\lambda_{2,0}+\epsilon}\right|$, where $\lambda_2$ is the second-smallest eigenvalue of the normalized graph Laplacian and $\epsilon=10^{-6}$.

Finally, we compare the computational complexity of the methods. Assuming that the input graph is sparse, i.e., $E=\mathcal{O}(N)$, the GR coarse-graining procedure scales linearly with the network size, $\mathcal{O}(N)$. However, GR first requires a hyperbolic embedding of the network using Mercator, whose standard time complexity is $\mathcal{O}(N^2)$. As shown in Appendix~\ref{apx:mercator}, GPU acceleration can substantially reduce the practical runtime of this embedding step. LRG requires $\mathcal{O}(N^3)$ time due to the eigendecomposition and matrix exponential operations. MagEdgePool and SpreadEdgePool have time complexity $\mathcal{O}(N^3)$ on sparse graphs when approximate diffusion distances are used.

\subsection{Computational analysis of Mercator on GPU}\label{apx:mercator}
Mercator~\cite{garcia2019mercator,jankowski2023d} is a reliable method for embedding complex networks into their underlying hyperbolic latent geometry. The algorithm combines machine-learning techniques with maximum-likelihood (ML) optimization to infer node coordinates in the hyperbolic disk by maximizing the agreement between the observed network topology and the geometric model. The original implementation has a time complexity that scales quadratically with the network size.

\begin{figure}[h]
	\centering
	\includegraphics[width=\linewidth]{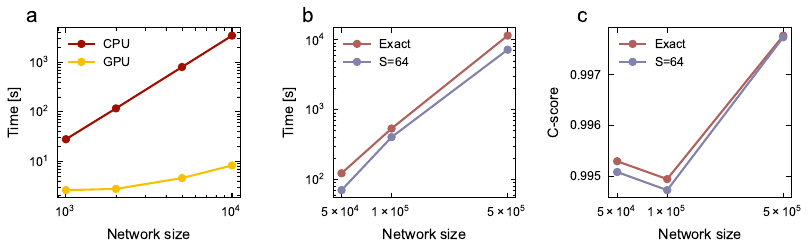}
    \caption{Computational performance of Mercator. (\textbf{a}) Embedding time as a function of network size for the CPU and GPU implementations. (\textbf{b}) Embedding time on the GPU using either exact maximum-likelihood (ML) calculations or negative sampling with $S=64$. (\textbf{c}) $c$-score between the original and inferred angular coordinates for embeddings obtained with exact ML and negative sampling. Model parameters are $\beta=2$, $\gamma=10$, and $\langle k \rangle=10$. All experiments were performed on a single H200 GPU.}
	\label{fig:mercator}
\end{figure}

In this work, we accelerate the ML stage by moving the most computationally expensive operations to a GPU. In Figure~\ref{fig:mercator}a, we report the time required to embed networks of different sizes. The GPU implementation, called \textit{cuMercator}, provides substantial speedup compared with the CPU version.
However, the asymptotic time complexity remains $\mathcal{O}(N^2)$. To reduce this cost, we introduce negative sampling into the ML calculation. Instead of computing the log-likelihood contribution from all non-neighbors at each newly proposed position, we sample only $S$ non-neighbors. This significantly reduces the cost of evaluating the non-neighbor contribution. In Figure~\ref{fig:mercator}b, we compare the embedding time obtained with exact ML calculations against the negative-sampling approximation with $S=64$. We observe a substantial reduction in runtime while retaining high-quality embeddings.
Negative sampling may prevent the final log-likelihood from reaching the exact optimum. Nevertheless, Figure~\ref{fig:mercator}c shows that the resulting $c$-score~\cite{muscoloni2017machine} remains very high, indicating strong agreement between the ordering of nodes in the original and inferred angular coordinates. 
These changes allow us to embed very large networks efficiently and use the inferred node coordinates for geometric renormalization.

\subsection{Choice of optimizer}\label{sec:optimizer}
In this work, we used Adam as the optimizer for GNN training. We also tested stochastic gradient descent (SGD) as a baseline. While SGD led to convergence on the synthetic networks, its performance was weaker on some real-world networks. Therefore, we used Adam in all experiments to ensure stable training and high accuracy across all considered networks.
It is worth noting one limitation of using Adam when analyzing training trajectories. With SGD on synthetic networks, we were able to extract raw gradients and directly compare the gradients of GNNs trained on the original and coarse-grained graphs. With Adam, however, the raw gradients were noisier and less stable for comparison. For this reason, we instead compared the softmax prediction trajectories, which provided a more stable basis for analyzing training alignment.

\subsection{Additional results}

\subsubsection{Synthetic networks}\label{apx:synth_networks}
\begin{figure}[h]
	\centering
	\includegraphics[width=\textwidth]{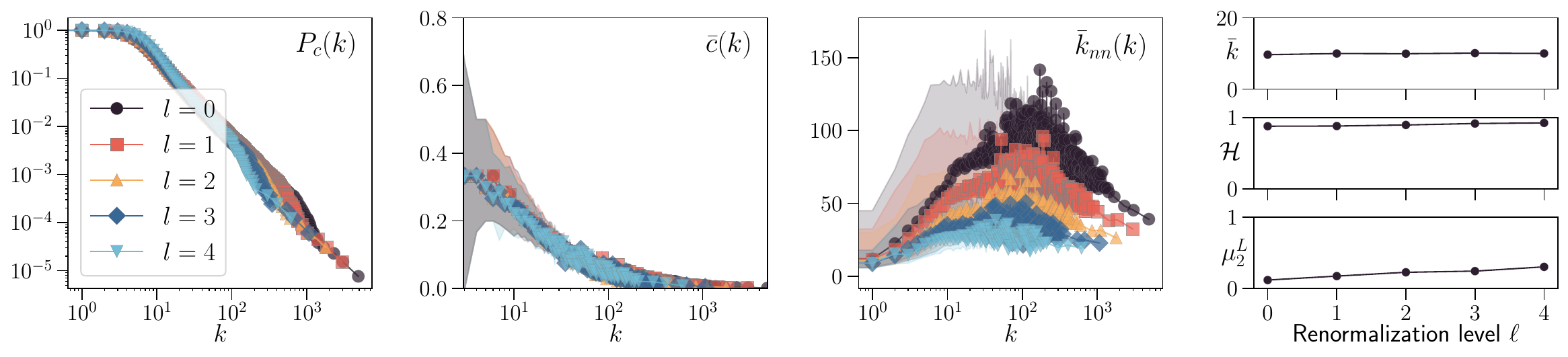}
    \caption{Topological validation of geometric renormalization in the ultra-small-world (USW). See the caption of Fig.~\ref{fig:GR} for panel details. Across all renormalization levels, these topological and mesoscale properties remain stable, confirming that geometric renormalization preserves the structural organization of the network in this regime as well. }
	\label{fig:usw_props}
\end{figure}

\begin{figure}[h]
    \centering
    \includegraphics[width=\textwidth]{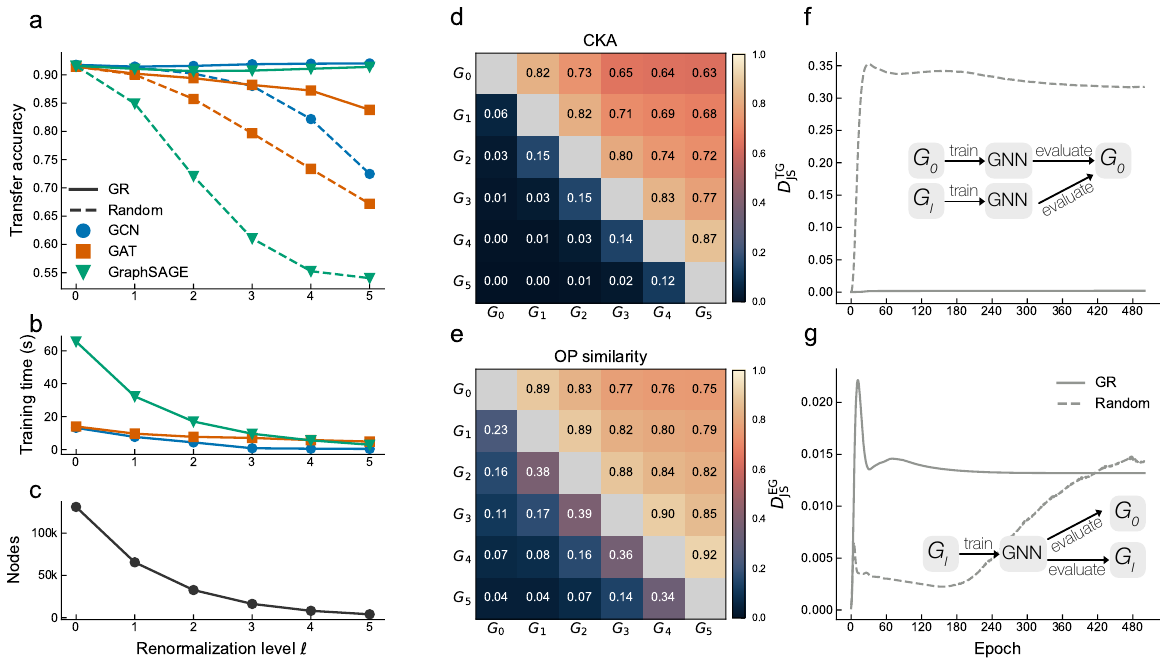}
    \caption{Zero-shot GNN transfer on synthetic networks in the USW regime. See the caption of Figure~\ref{fig:synthetic_panel} for more details.}
	\label{fig:usw}
\end{figure}

Figure~\ref{fig:usw_props} shows the topological validation of geometric renormalization in the ultra-small-world (USW) regime. Similarly to the small-world regime, the topological properties are preserved. Figure~\ref{fig:usw} summarizes the zero-shot transfer results.

\newpage
\subsubsection{Real networks}
\label{apx:real_networks}
\begin{figure}[htbp]
    \centering
    \includegraphics[width=\textwidth]{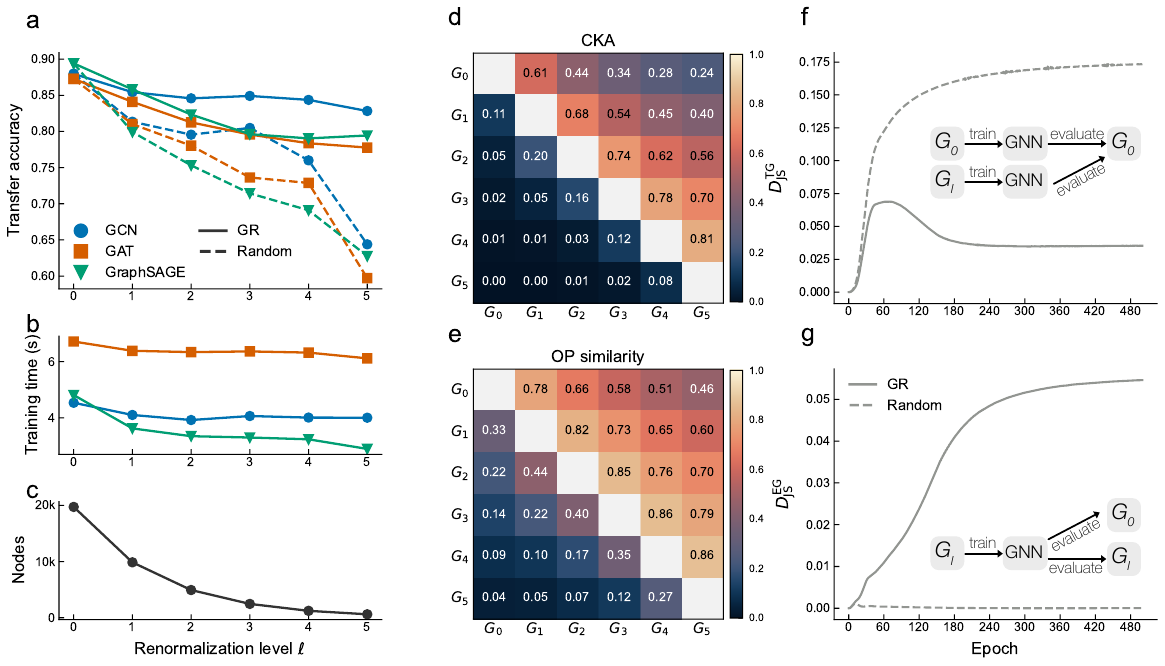}
    \caption{\textbf{Zero-shot GNN transfer on PubMed dataset}. See the caption of Fig.~\ref{fig:synthetic_panel} for more detail.} 
	\label{fig:pubmed_vs_random}
\end{figure}

\begin{figure}[htbp]
    \centering
    \includegraphics[width=\textwidth]{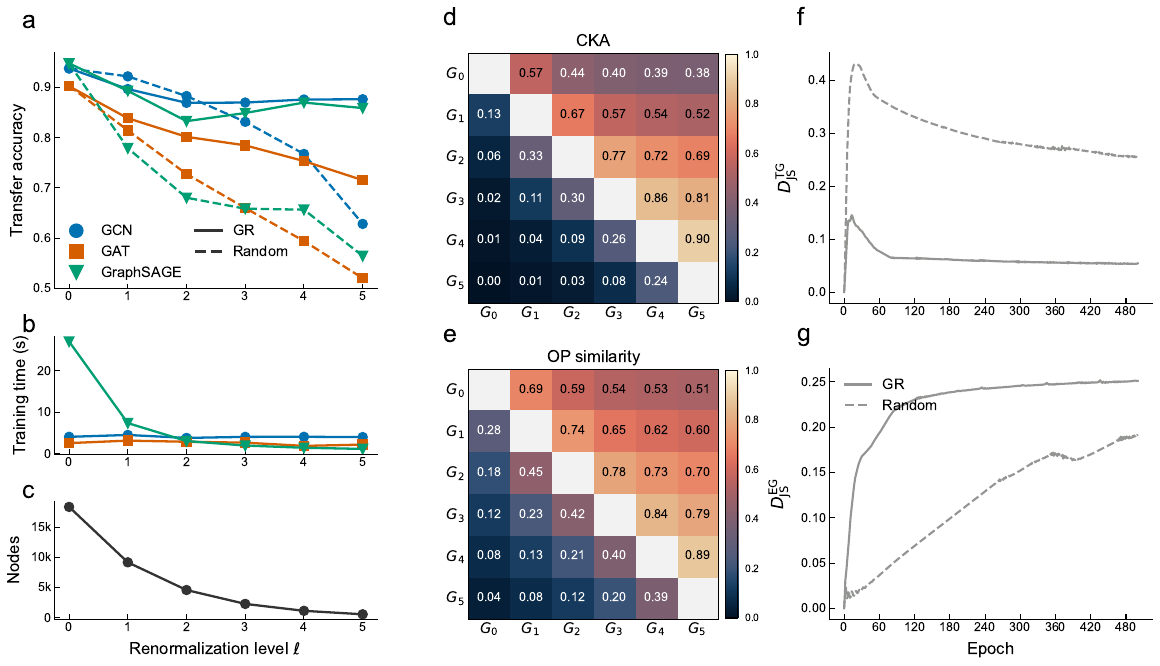}
    \caption{\textbf{Zero-shot GNN transfer on CS dataset}. See the caption of Fig.~\ref{fig:synthetic_panel} for more detail.} 
	\label{fig:cs_vs_random}
\end{figure}

\begin{figure}[htbp]
    \centering
    \includegraphics[width=\textwidth]{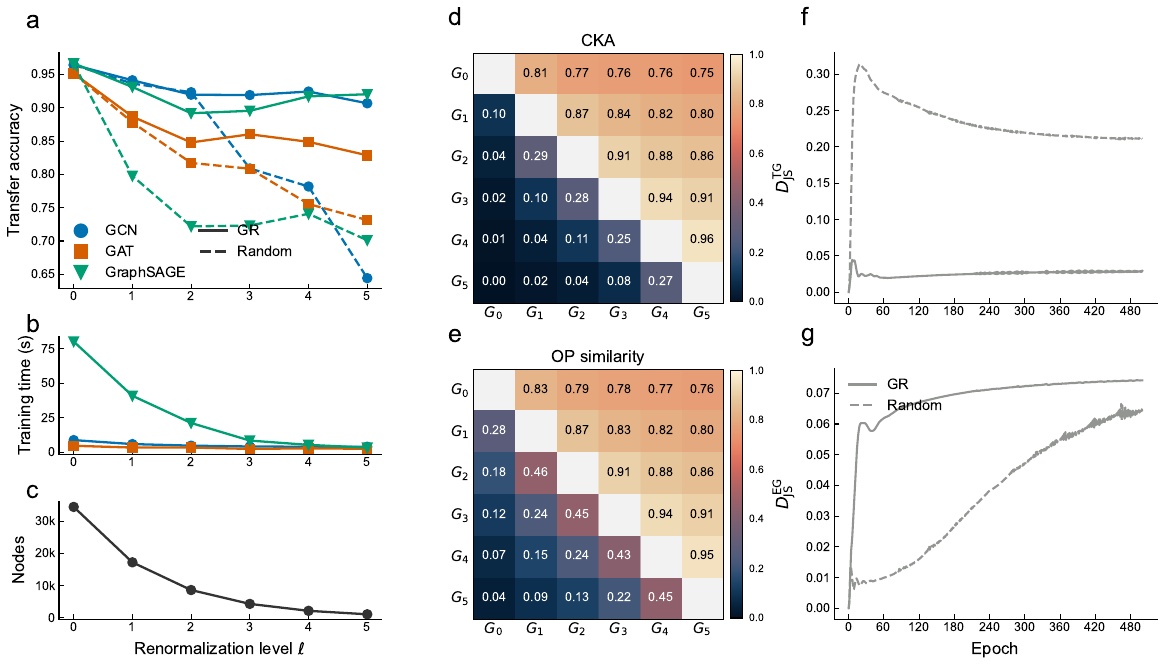}
    \caption{\textbf{Zero-shot GNN transfer on Physics dataset}. See the caption of Fig.~\ref{fig:synthetic_panel} for more detail.} 
	\label{fig:physics_vs_random}
\end{figure}

\begin{figure}[htbp]
    \centering
    \includegraphics[width=\textwidth]{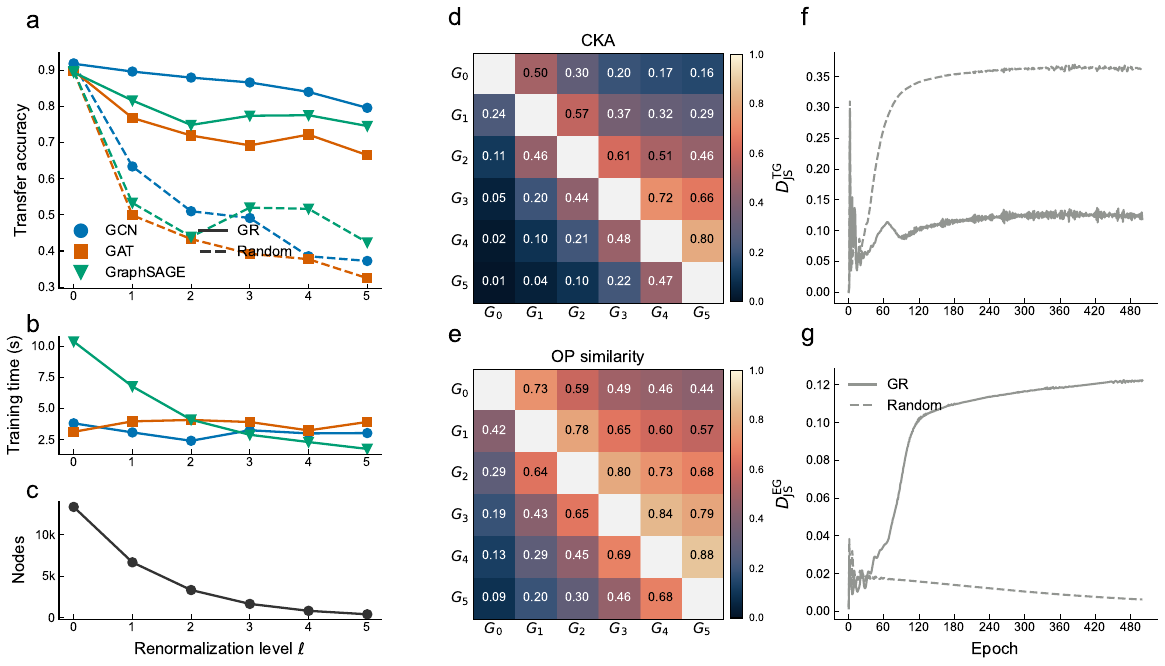}
    \caption{\textbf{Zero-shot GNN transfer on Computers dataset}. See the caption of Fig.~\ref{fig:synthetic_panel} for more detail.} 
	\label{fig:computers_vs_random}
\end{figure}

\begin{figure}[htbp]
    \centering
    \includegraphics[width=\textwidth]{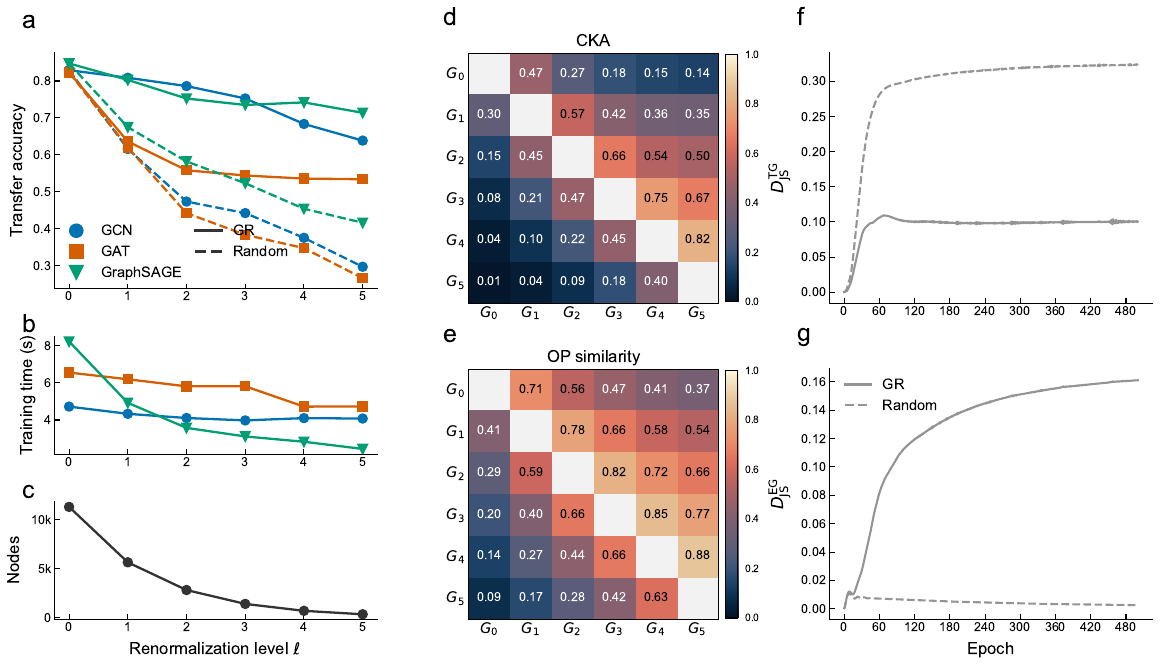}
    \caption{\textbf{Zero-shot GNN transfer on WikiCS dataset}. See the caption of Fig.~\ref{fig:synthetic_panel} for more detail.} 
	\label{fig:wikics_vs_random}
\end{figure}

\begin{figure}[htbp]
    \centering
    \includegraphics[width=\textwidth]{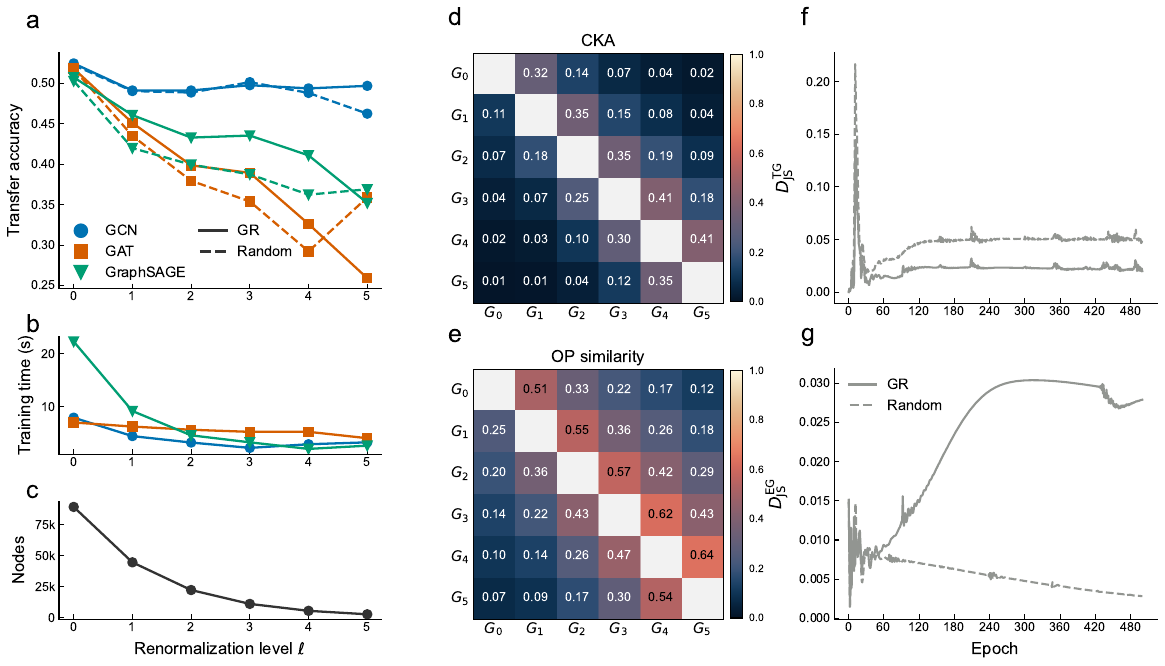}
    \caption{\textbf{Zero-shot GNN transfer on Flickr dataset}. See the caption of Fig.~\ref{fig:synthetic_panel} for more detail.} 
	\label{fig:flickr_vs_random}
\end{figure}

\begin{figure}[htbp]
    \centering
    \includegraphics[width=\textwidth]{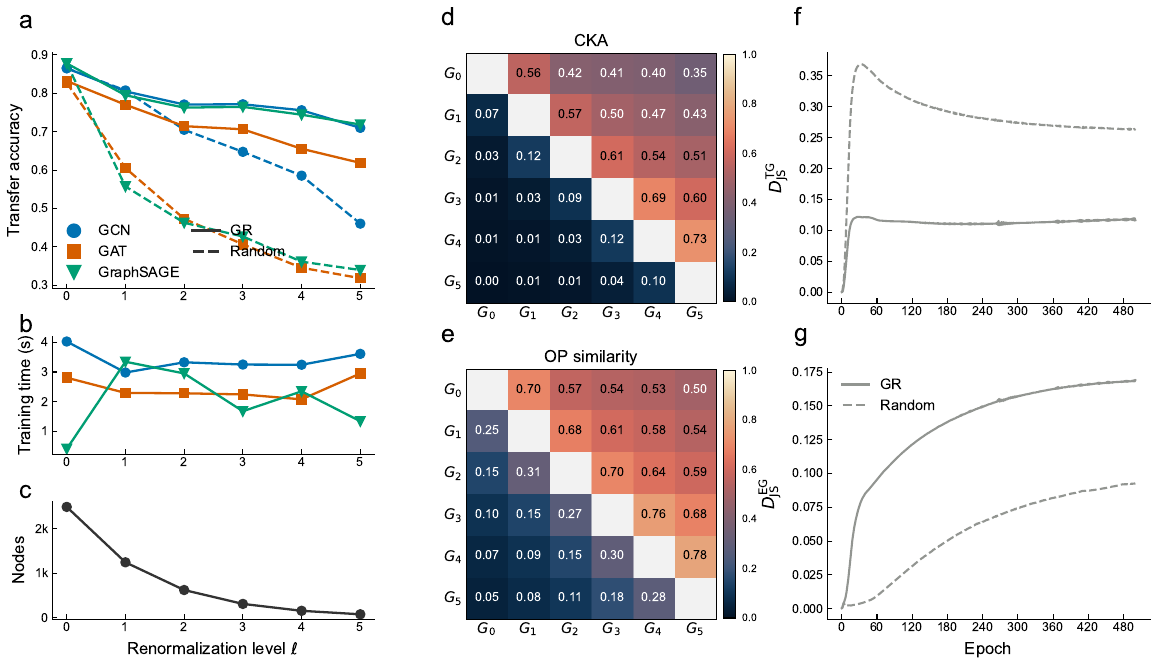}
    \caption{\textbf{Zero-shot GNN transfer on Cora dataset}. See the caption of Fig.~\ref{fig:synthetic_panel} for more detail.} 
	\label{fig:cora_vs_random}
\end{figure}

Figures~\ref{fig:pubmed_vs_random}--\ref{fig:cora_vs_random} show transfer accuracy and 
training time across renormalization levels for the remaining real-world datasets.

\newpage

\newpage

\newpage

\subsubsection{Comparison with Random-T and Random-G baselines}
\label{apx:random_variants}

In the main text, we compared geometric renormalization (GR) against a fully randomized baseline (\textit{Random}), in which both the network topology and the geometric coordinates used to merge nodes are randomized before coarse-graining. To disentangle the individual contribution of these two factors, we here report results for two partial baselines: \textit{Random-T}, in which only the topology is randomized while node coordinates are kept as inferred by Mercator, and \textit{Random-G}, in which only the geometric coordinates are randomized while the original topology is preserved. Comparing GR against these two variants isolates whether preserving topology, preserving geometry, or their combination is responsible for the transfer performance observed in the main text.

\begin{figure}[htbp]
    \centering
    \includegraphics[width=\textwidth]{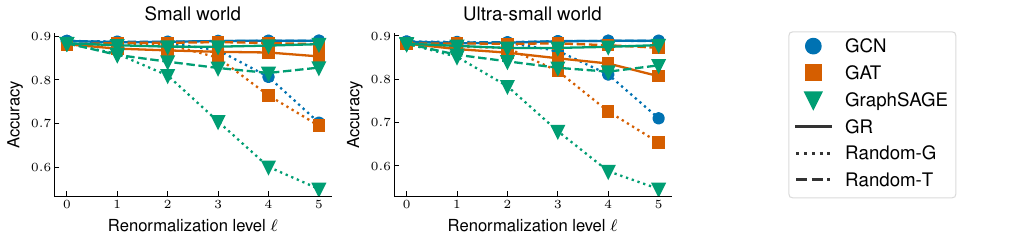}
    \caption{\textbf{Zero-shot GNN transfer accuracy for SW and USW regimes, against Random-T and Random-G renormalization methods}. See the caption of Fig.~\ref{fig:synthetic_panel} for details on models and renormalization methods shown.}
	\label{fig:synth_vs_random-gt}
\end{figure}

Figure~\ref{fig:synth_vs_random-gt} shows the zero-shot transfer accuracy for the synthetic networks in both the ultra-small-world (USW) and small-world (SW) regimes, comparing GR against Random-T and Random-G. 

\begin{figure}[htbp]
    \centering
    \includegraphics[width=\textwidth]{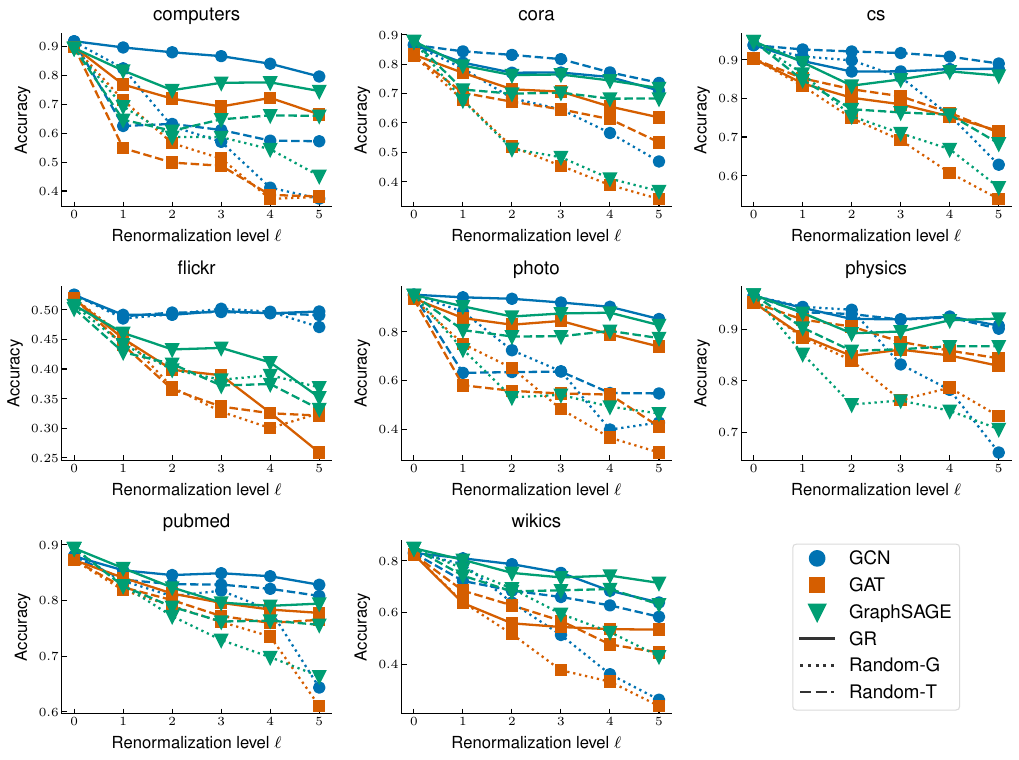}
    \caption{\textbf{Zero-shot GNN transfer accuracy across all real-world datasets, against Random-T and Random-G renormalization methods}. Each panel corresponds to one dataset. See the caption of Fig.~\ref{fig:synthetic_panel} for details on models and renormalization methods shown.}
	\label{fig:panel_vs_random-gt}
\end{figure}

Figure~\ref{fig:panel_vs_random-gt} shows the zero-shot transfer accuracy for each of the eight real-world datasets, comparing GR against Random-T and Random-G.

Interestingly, the three architectures respond differently to the Random-T baseline. We hypothesize that this reflects how each model aggregates neighborhood information. One possible explanation is that each architecture aggregates neighborhood information differently: GraphSAGE separates self and neighbor transformations, GCN applies a shared transformation across both, and GAT learns attention weights over neighbors, which may make them differently sensitive to randomized topology.

\newpage
\subsubsection{Topological validation}\label{apx:realnets_props}

\begin{figure}[h]
	\centering
	\includegraphics[width=\linewidth]{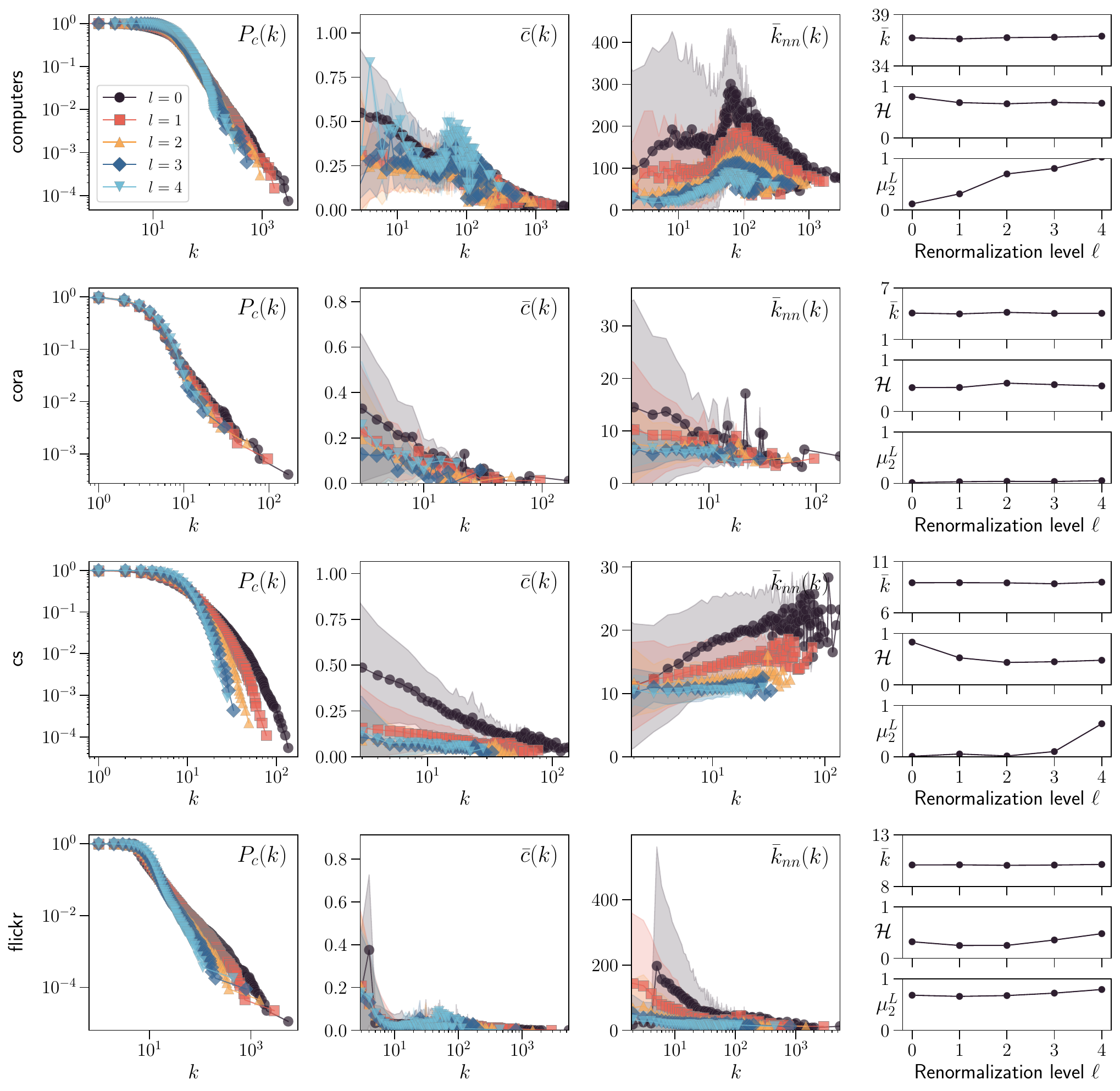}
	\caption{Topological validation of GR for the real-world datasets considered (part 1 of 2).}
	\label{fig:all_datasets_gr_props1}
\end{figure}

\begin{figure}[h]
	\centering
	\includegraphics[width=\linewidth]{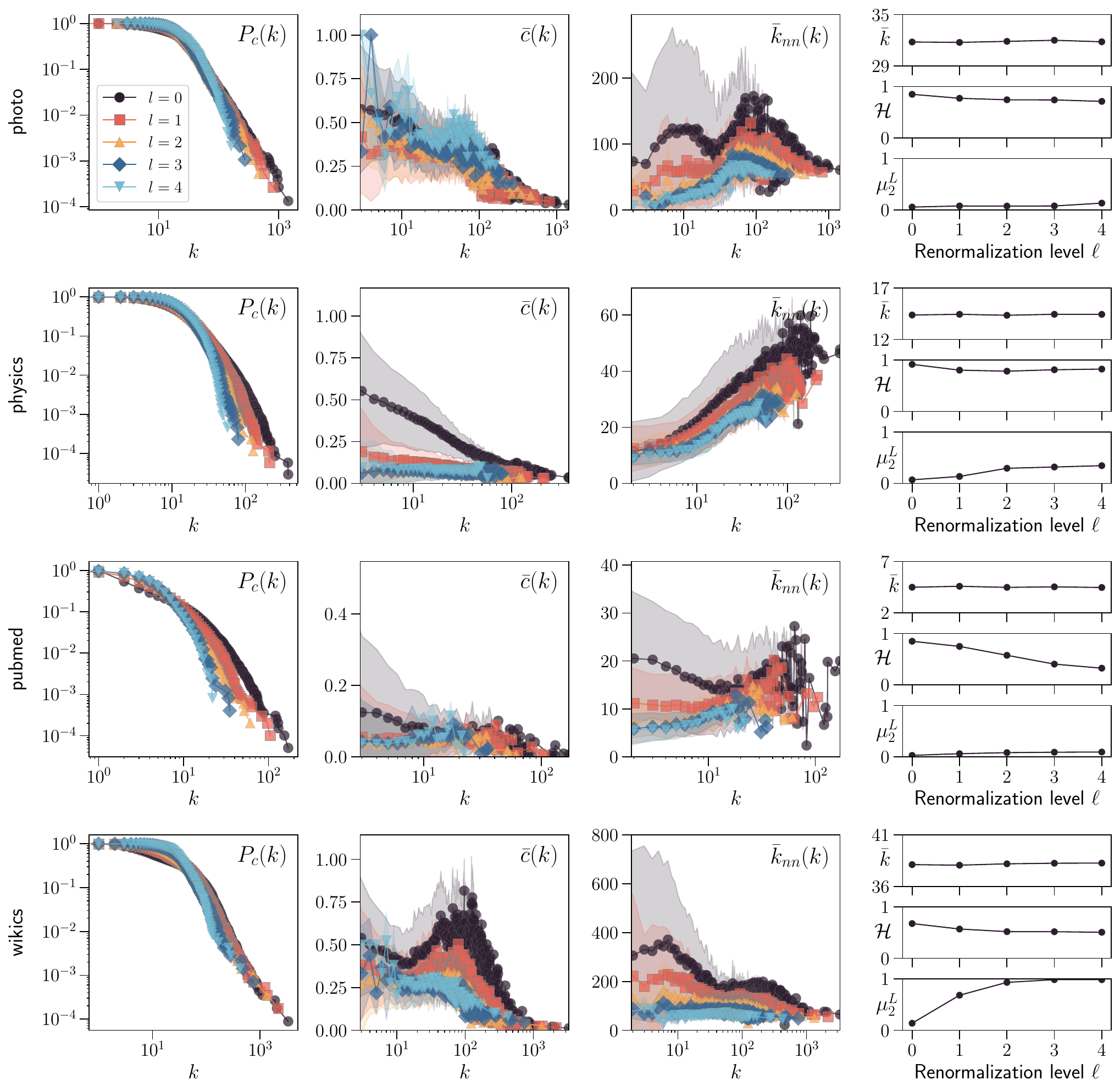}
	\caption{Topological validation of GR for the real-world datasets considered (part 2 of 2).}
	\label{fig:all_datasets_gr_props2}
\end{figure}

Figures~\ref{fig:all_datasets_gr_props1} and~\ref{fig:all_datasets_gr_props2} show the topological validation of GR across all real-world networks considered. 

\begin{figure}[h]
	\centering
	\includegraphics[width=\linewidth]{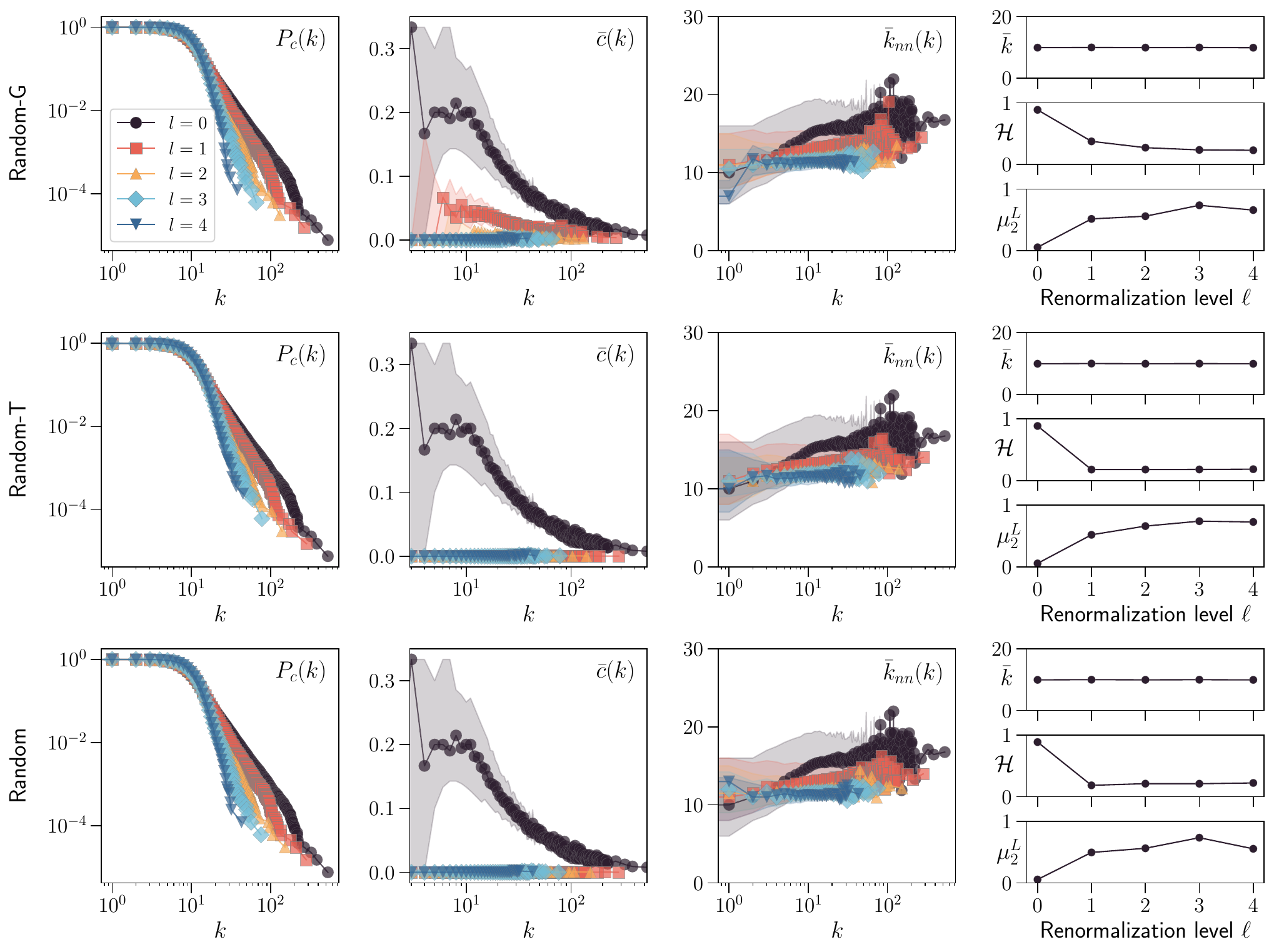}
	\caption{Topological validation of the SW regime under three random variants of the renormalization method.}
	\label{fig:methods_panel_props}
\end{figure}

\begin{figure}[h]
	\centering
	\includegraphics[width=\linewidth]{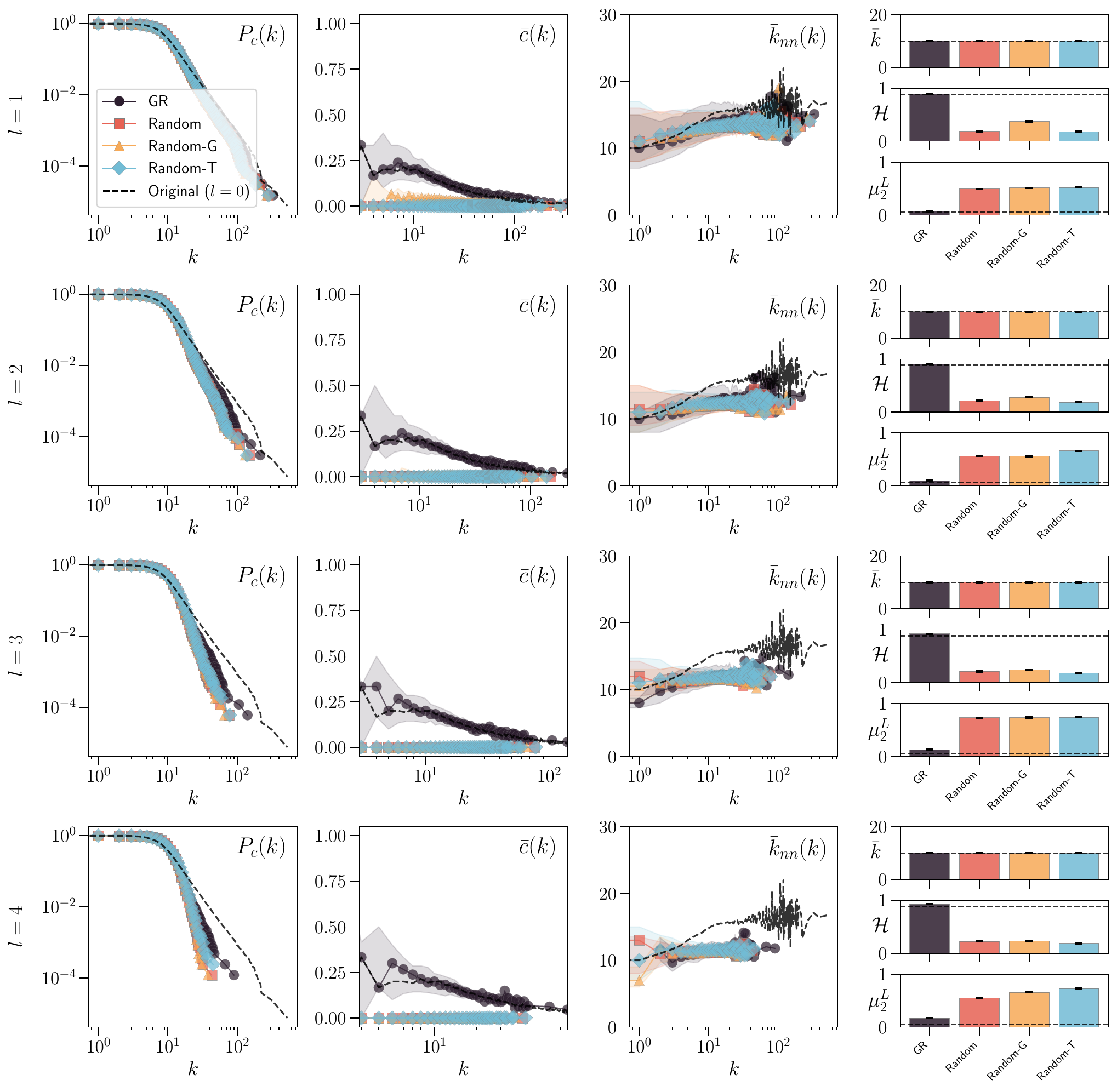}
	\caption{Topological validation of the SW regime under GR and three random variants-Random, Random-G, and Random-T of the renormalization method, shown for renormalization levels $l=1,2,3,4$. Dashed black lines indicate the reference values of the original (non-renormalized) network at $l=0$.}
	\label{fig:methods_panel_levels}
\end{figure}

Figures~\ref{fig:methods_panel_props} and~\ref{fig:methods_panel_levels} further show the topological validation of the SW regime under the random renormalization variants (Random, Random-T, and Random-G), the latter broken down by renormalization level $l=1,2,3,4$.

\end{document}